%% file: acl_latex.tex
\pdfoutput=1

\documentclass[11pt]{article}

\usepackage{acl}
\usepackage{tcolorbox}

\usepackage[ruled,linesnumbered]{algorithm2e}
\usepackage{times}
\usepackage[]{tcolorbox}
\usepackage{latexsym}
\usepackage{amsmath}
\usepackage{physics}
\usepackage{bbding}
\usepackage{pifont}
\usepackage{wasysym}
\usepackage{amssymb}
\usepackage[T1]{fontenc}
\usepackage{placeins}
\usepackage[utf8]{inputenc}

\usepackage{microtype}

\usepackage{inconsolata}

\usepackage{graphicx}
\newcommand{\cmark}{\ding{51}}%
\newcommand{\xmark}{\ding{55}}%
\usepackage{xcolor}
\usepackage{soul}
\usepackage{float}
\usepackage{times}
\usepackage{latexsym}
\usepackage{multirow}
\usepackage{graphicx}
\usepackage{makecell}
\usepackage{booktabs}
\usepackage{colortbl}
\definecolor{aqua}{rgb}{0.0, 1.0, 1.0}
\definecolor{hot pink}{rgb}{1.0, 0.31, 0.61}

\usepackage{xcolor,pifont}
\newcommand*\colourcheck[1]{%
  \expandafter\newcommand\csname #1check\endcsname{\textcolor{#1}{\ding{52}}}%
}
\title{SLoW: Select Low-frequency Words! Automatic Dictionary Selection\\for Translation on Large Language Models}

\author{
    Hongyuan Lu$^{\heartsuit\clubsuit}$\Thanks{\hspace{1mm}Equal Contribution.}, Zixuan Li
$^{\spadesuit*}$, Zefan Zhang$^\diamond$, Wai Lam$^\heartsuit$\\
    $\heartsuit$The Chinese University of Hong Kong\\
    $^\spadesuit$Cyber Science and Engineering, Southeast University\\
    $^\clubsuit$FaceMind Corporation\\
    $^\diamond$ College of Computer Science and Technology, Jilin University\\
    \{hylu,wlam\}@se.cuhk.edu.hk, zixuan.li@seu.edu.cn, zefan23@mails.jlu.edu.cn
}

\begin{document}
\maketitle
\begin{abstract}
 There are more than 7,000 languages around the world, and current Large Language Models (LLMs) only support hundreds of languages. Dictionary-based prompting methods can enhance translation on them, but most methods use all the available dictionaries, which could be expensive. Instead, it will be flexible to have a trade-off between token consumption and translation performance. This paper proposes a novel task called \textbf{A}utomatic \textbf{D}ictionary \textbf{S}election (\textbf{ADS}). The goal of the task is to automatically select which dictionary to use to enhance translation. We propose a novel and effective method which we call \textbf{S}elect \textbf{Lo}w-frequency \textbf{W}ords! (\textbf{SLoW}) which selects those dictionaries that have a lower frequency. Our methods have unique advantages. First, there is no need for access to the training data for frequency estimation (which is usually unavailable). Second, it inherits the advantage of dictionary-based methods, where no additional tuning is required on LLMs. Experimental results on 100 languages from FLORES indicate that SLoW surpasses strong baselines, and it can obviously save token usage, with many languages even surpassing the translation performance of the full dictionary baseline.\footnote{A shocking fact is that there is no need to use the actual training data (often unobtainable) for frequency estimation,  and an estimation frequency obtained using public resources is still apparently effective in improving translation with ChatGPT and LLaMa, and DeepSeek.}\footnote{\url{https://github.com/HongyuanLuke/SLoW}.}
\end{abstract}

 \section{Introduction}
Large Language Models (LLMs) have exhibited many exciting capabilities such as chain-of-thought reasoning \citep{wang-etal-2023-towards,10.5555/3600270.3602070}, neural machine translation \citep{2023arXiv230506575L,lu-etal-2023-trip,tang-etal-2024-metrics,zhu-etal-2024-multilingual,zhu-etal-2024-clean,lu-etal-2024-revamping}, code understanding and code generation \citep{li-etal-2023-codeie,zhang-etal-2023-self,hou-etal-2025-lne}, dialogue generation \citep{lu-etal-2022-partner, lu-lam-2023-pcc}, and even spatial reasoning \citep{hu2024chainofsymbol}. While LLMs have demonstrated their exciting performances on a wide range of tasks, they are usually English-centric, and their multilingual abilities are usually limited, especially in those low-resourced languages. Dictionary-based methods effectively improve multilingual capabilities by adding word mappings into the prompt \citep{lu-etal-2024-chain, 2024arXiv241101141L}. Yet, most current dictionary-based translation methods for LLMs use all the matching dictionaries greedily, and there are so far no systematic guidelines or architecture to select which dictionaries to use. Such a greedy strategy can lead to unnecessary token consumption, as LLMs may have no problems understanding some of the words. Furthermore, too much irrelevant or redundant information can distract LLMs \citep{10.5555/3618408.3619699}. Therefore, we propose a novel Natural Language Processing task called \textbf{A}utomatic \textbf{D}ictionary \textbf{S}election (\textbf{ADS}). The input into the task of ADS is a set of available dictionaries and a set of input translation source instances. The goal of ADS is to maximise the translation performance by using only a subset of the dictionaries, so there can be a trade-off where more dictionaries to be used may have a better translation performance. Therefore, we constrain ADS to use no more than a certain number of words, $\mathcal{W}$ words, and in this paper, we make it the method with the lowest number of dictionaries among the methods in comparison. \par
To tackle ADS, we propose a novel and effective method which we call \textbf{S}elect \textbf{Lo}w-frequency \textbf{W}ords! (\textbf{SLoW}). SLoW selects the dictionaries that have a lower frequency in the training data. We postulate that this is because how frequently the words are presented in the training data is directly related to how well the LLMs understand them, and adding the dictionary of those low-frequency words makes it easier for LLMs to understand and translate less frequent and less well-learned words.
\par 
Further analysis indicates that such methods are better than many competitive baselines such as using nouns, verbs, adjectives, or their combination greedily. Surprisingly, we also found that selecting partial dictionaries with SLoW can even beat full dictionary usage in some cases. This paves a new research direction to optimize the selection of dictionary usage automatically for dictionary-based translation methods on LLMs.
\par 
We emphasize the shocking fact that there is no need to obtain the actual training data, which is often unobtainable, and online public resources can be used to improve LLaMa and ChatGPT through SLoW. This suggests a good estimation of online resources on word frequencies in the training data from ChatGPT and LLaMa.
\par 
Our contributions are three-fold:
\begin{itemize}
\setlength\itemsep{0em}
    \item We propose a novel task called \textbf{A}utomatic \textbf{D}ictionary \textbf{S}election, where it considers the trade-off between dictionaries and performance to be used when prompting LLMs.
    \item We propose a novel method to tackle ADS, which we call \textbf{S}elect \textbf{Lo}w-frequency \textbf{W}ords! (\textbf{SLoW}). SLoW selects the dictionaries that have a lower frequency in the training data.
    \item We conduct experiments on 100 languages from FLORES for Machine Translation. Experimental results indicate that SLoW beats competitive baselines and can even surpass the case when full dictionaries are used.
 
\end{itemize}
\section{Prior Work}
\paragraph{Neural Machine Translation via LLMs} Research on effective methods for prompting English-centric Large Language Models (LLMs) for non-English tasks, including standard cross-lingual tasks like Multilingual Neural Machine Translation (MNMT), remains limited. Most existing studies have primarily focused on evaluating the translation performance of English-centric LLMs using prompts such as `Translate to \{language\_name\}: {text}' \citep{NEURIPS2020_1457c0d6,2021arXiv211210668L,2022arXiv221105100W,2022arXiv220501068Z}. Various prompt formats have also been explored \citep{10.1145/3411763.3451760,2023arXiv230304048W}. Additionally, \citet{2022arXiv220211822G} examined the use of prompts to regulate aspects like formality or specific dialects in a generation. Furthermore, \citet{2022arXiv221202437A} and \citet{2022arXiv221109102V} investigated selecting appropriate in-context examples to enhance the machine translation quality of LLMs. Generally speaking, the research of MNMT has now scaled to hundreds of languages as seen with FLORES \citep{nllb2022}.
\paragraph{Dictionary-based Method for Neural Machine Translation} 
This research is closely tied to the concept of lexical constraints in machine translation, which can be categorized into hard constraints \citep{hokamp-liu-2017-lexically,post-vilar-2018-fast} and soft constraints \citep{2019arXiv190409107S,dinu-etal-2019-training,10.5555/3491440.3491936}.
\par
Several studies have investigated the use of dictionaries in supervised machine translation. For instance, \citet{2016arXiv161007272Z} enhanced neural machine translation (NMT) by incorporating a bilingual dictionary to include rare or unseen words absent from the bilingual training data. Similarly, \citet{arthur-etal-2016-incorporating} improved the translation of rare words by integrating discrete translation lexicons and using the attention vector to estimate relevant lexical probabilities. \citet{10.1145/3377713.3377801} leveraged dictionaries to generate synthetic parallel data, enhancing NMT training. \citet{lu-etal-2024-chain} used chained dictionaries to enhance machine translation with LLMs by leveraging intermediate auxiliary languages.
\par
While much of the prior work has centred on using dictionaries for machine translation tasks, how to effectively select a subset of dictionaries to achieve a good trade-off remains unexplored. In contrast, ADS is the first task that considers which types of dictionaries should be used on LLMs for automatic machine translation.
\par 
Later in 2026, \citet{2026arXiv260402176L} proposed to use sentence-level textual frequency to enhance machine translation.

\section{Automatic Dictionary Selection}
\subsection{Translation with LLMs} 
We start by introducing our proposed task, namely automatic dictionary selection. The goal of such a task is to select appropriate dictionaries in order to maximise the performance of the succeeding generation task by adding the dictionaries into the prompt, and this paper focuses on the setting of neural machine translation on LLMs which use dictionaries for translation \citep{lu-etal-2024-chain}.
\par 
LLM can be regarded as a Seq2Seq neural network \citep{seq2seq} to translate an input language into the output language while maintaining the semantical equivalence and maximise the following likelihood:
\begin{align*}
    P\,(\vb*{\hat{t}}\mid \vb*{i}, \vb*{s}, \vb*{d})=\prod_{j=1}^{\mathbb{T}}P\,(\hat{t}_j\mid \hat{t}_1,..., \hat{t}_{j-1}, \vb*{i}, \vb*{s}, \vb*{d}),
\end{align*}
where $\mathbb{T}$ represents the length of the generated translation output and $\hat{t_j}$ represents the word at the position $j$ that has been inferenced. $\vb*{s}$ represents the source sentences, $\vb*{d}$ represents the dictionaries that has been selected to be used for improving the translation. $\vb*{i}$ represents the translation instruction to guide the LLMs to translate the words. A typical translation instruction could be:
\begin{tcolorbox}
Translate the following sentence from <source language> into <target language>: <source sentence>
\end{tcolorbox}
\par
\subsection{Automatic Dictionary Selection}
However, which dictionaries to be used $\vb*{d}$ has not been explored to our best knowledge. That means, in previous works, all the dictionaries are provided and inserted into the prompt as long as there is a match regardless of how useful they will be. However, intuitively speaking, this could not be the best choice. Even if the results are not maximised, one might want to reduce the computational cost as a trade-off to gain limited improvement with dictionary methods. Therefore, we propose a novel task ADS to automatically select dictionaries. The task is formulated as:
\begin{align*}
    \hat{\mathcal{D}}=\mathcal{M}(\mathcal{D}, \mathcal{L}),
\end{align*}
where $\mathcal{M}$ is a selection function, where we select a subset of dictionary $\hat{\mathcal{D}}$ from the complete dictionary $\mathcal{D}$ for succeeding downstream task dataset $\mathcal{L}$. Since such a selection might always be maximised by selecting the full dictionary, we define a dictionary size $\mathcal{V}$ which is usually lower than the full dictionary size, and the goal of ADS is to find a better function $\mathcal{M}$ that maximises the final performance on the $\mathcal{L}$ with a subset of the dictionary, namely, $\hat{\mathcal{D}}$, which has a dictionary size of $\mathcal{V}$.
\par 
\subsection{Select Low-frequency Words!}
In this paper, we propose a novel and effective method which we call \textbf{S}elect \textbf{Lo}w-frequency \textbf{W}ords! (\textbf{SLoW}). SLoW selects the dictionaries that have a lower frequency in the training data: 

\par
\begin{equation}
\label{diff}
     \hat{\mathcal{D}}=\mathrm{first}(\mathrm{sort}_{\bar{x}_i\in \mathcal{D}}(\mathcal{G}(\bar{x}_i, \mathcal{T})), \mathcal{V}),
\end{equation}
where $\mathrm{first}$ selects from a sorted list in acending order created from $\mathrm{sort}$ to get the $\mathcal{V}$ lowest-frequency dictionaries selected by a frequency estimation function $\mathcal{G}$ with the training set $\mathcal{T}$ used for training the LLMs. Note that here for the translation task in this paper, English frequency can used as a standard, because most LLMs are English-centric.
\par
We surprisingly found its usefulness despite it being simple, compared to various strong baselines that we have compared. This is yet intuitively aligned with our expectation, as we definitely would like to enhance LLMs' knowledge if that part of knowledge is not trained well. Data scarcity, i.e., low-frequency is a common reason for not training that part of the knowledge well.
\par
In this paper, we have attempted various baselines. Since we conduct experiments on hundreds of languages from FLORES, and our computational resources are limited, we explore the setting where we set a fixed $\mathcal{V}$.
\section{Experimental Setup}
\subsection{Datasets and Evaluation Metrics}
We evaluate the task of Neural Machine Translation with the dictionary-based setting where dictionaries are used to improve machine translation \citep{lu-etal-2024-chain}. Under this setting, low-resourced languages play an important role, because dictionary-based methods are particularly useful on them \citep{lu-etal-2024-chain}. A very useful dataset is FLORES \citep{nllb2022}, where we use 100 languages from FLORES devtest. This dataset comprises 1,012 sentences sourced from English Wikipedia, spanning diverse topics and domains (we randomly sample 200 instances). These sentences have been meticulously translated into hundreds of languages by professional translators. Since they are professionally translated by human experts into parallel languages, it is suitable for our use.
\par 
For the evaluation metrics, we report the chrF \citep{popovic-2015-chrf} and the BLEU \citep{papineni-etal-2002-bleu} evaluations provided by the sacreBLEU repository.\footnote{\url{https://github.com/mjpost/sacrebleu}} We also use evaluate with COMET scores using wmt22-comet-da\footnote{\url{https://github.com/Unbabel/COMET}} \citep{rei-etal-2020-comet} across all the experiments.
\par
For space reasons, we present the language class of our experiments for XX translation in Table \ref{lcc} and Table \ref{lcc2} in the Appendix.
\begin{figure*}[t!]
\begin{center}
\vspace{0mm}
\centerline{
\includegraphics[width=16cm]{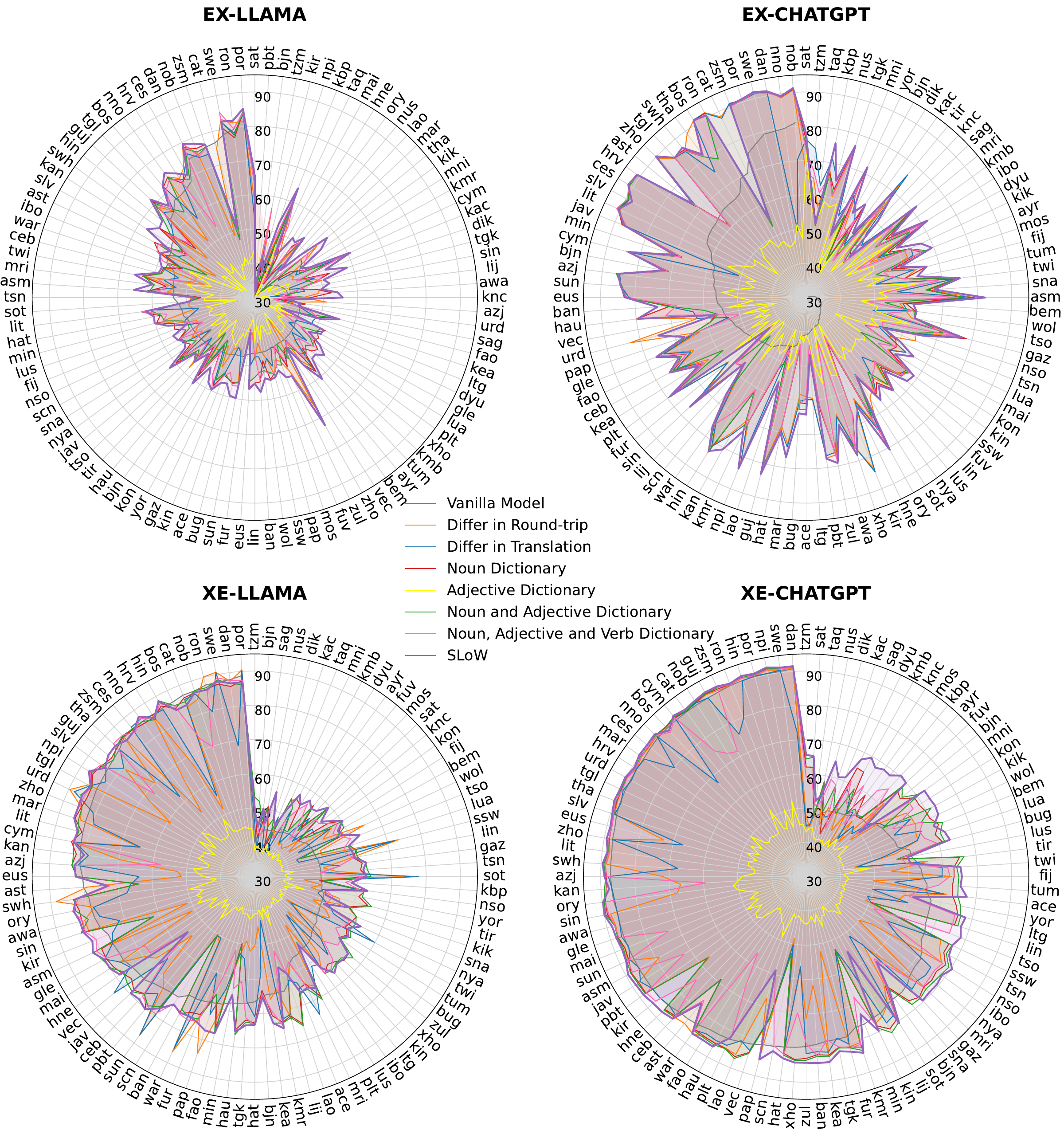}}
    \caption{Performance of LLaMa and ChatGPT in COMET scores on the task of Machine Translation both into English and from English translation on FLORES with different ADS methods. The top-left one is the translation from English on LLaMa, the top-right is the translation from English on ChatGPT, the bottom-left is the translation to English on LLaMa,  and the bottom-right is the translation to English on LLaMa. It is obvious that our proposed method SLoW is the best, surpassing many strong baselines. Such a phenomenon can be consistently observed across many low-resourced and high-resourced languages, demonstrating the effectiveness of our methods. For space reasons, more results on BLEU, chrF, evaluations and on DeepSeek-V3 in the Appendix in Table \ref{summemnlp}.}
    \label{fig:all}
\end{center}
\vspace{-5mm}
\end{figure*}
\begin{figure*}[t!]
\begin{center}
\vspace{0mm}
\centerline{
\includegraphics[width=16cm]{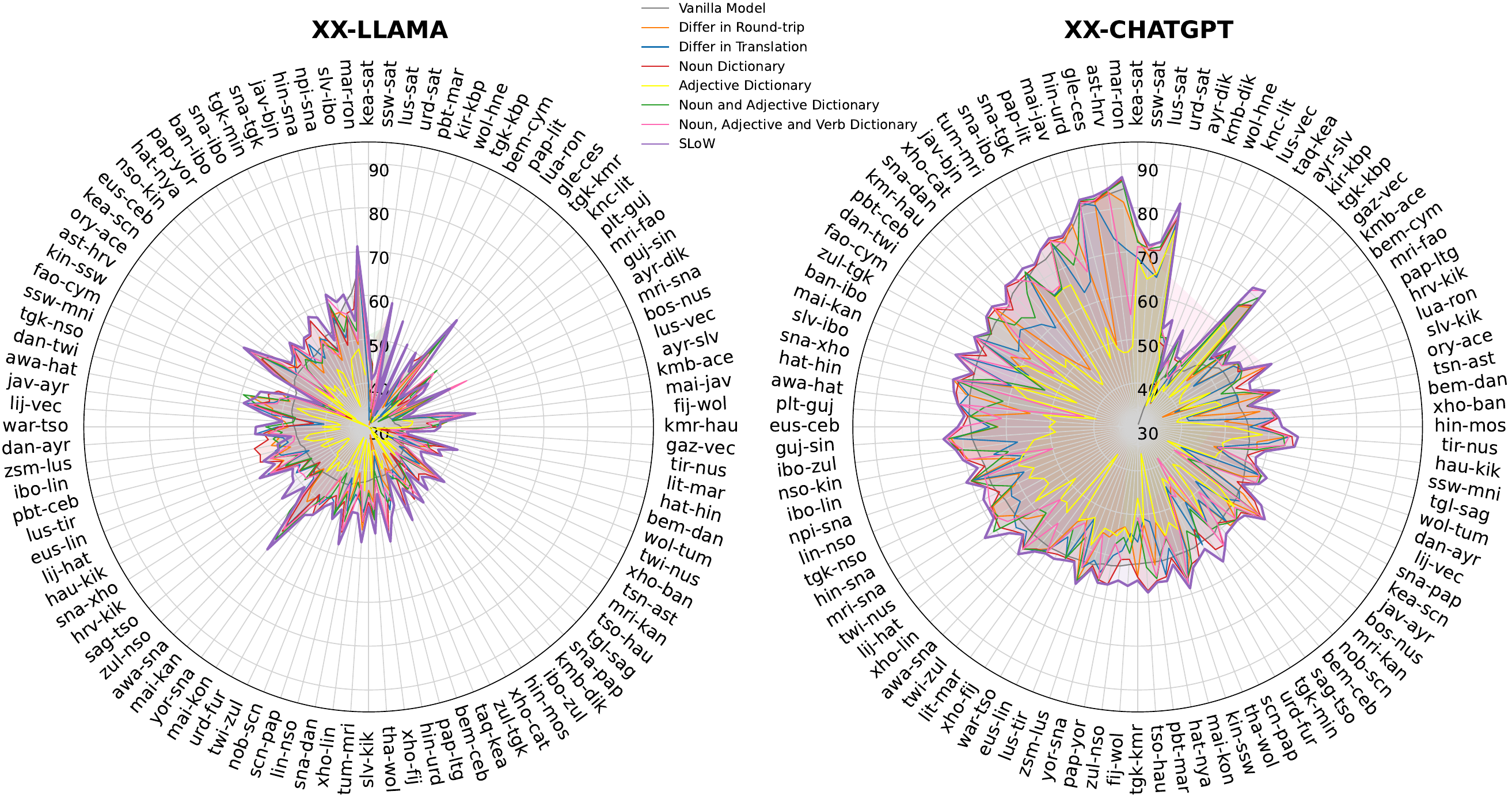}}
    \caption{Performance of LLaMa and ChatGPT the task of Machine Translation on non-English-centric translation in COMET scores on non-English-centric translation on FLORES with different ADS methods. It is obvious that our proposed method SLoW is the best, surpassing many strong baselines. Such a phenomenon can be consistently observed across many translation pairs, demonstrating the effectiveness of our methods. For space reasons, more results on BLEU, chrF, evaluations and on DeepSeek-V3 in the Appendix in Table \ref{summemnlp}.}
    \label{fig:xx}
\end{center}
\vspace{-5mm}
\end{figure*}
\begin{table*}
\centering
    \setlength\tabcolsep{1.5pt}
    \setlength\extrarowheight{0pt}
\begin{tabular}{l|cccc|ccc}
\hline
\noalign{\vskip 1mm}  
\textbf{Direction} & \textbf{\# improved} & \textbf{> 5 points} & \textbf{> 10 points} & \textbf{> 20 points} & \textbf{\# degraded} & \textbf{> 5 points} & \textbf{> 20 points}\\
\noalign{\vskip 1mm}  
\hline
\noalign{\vskip 1mm} 
En-X-LLaMa & 88/100 & 65/88 & 50/88 & 22/88 & 12/100 & 4/12 & 3/12 \\  
En-X-CHATGPT\,\, & 75/100 & 36/75 & 15/75 & 1/75 & 25/100 & 10/25 & 4/25  \\  
X-En-LLaMa & 76/100 & 63/76 & 57/76 & 39/76 & 24/100 & 10/24 & 5/24  \\  
X-En-CHATGPT\,\, & 92/100 & 50/92 & 43/92 & 33/92 & 8/100 & 5/8 & 2/8  \\ 
X-X-LLaMa & 93/100 & 46/93 & 7/93 & 1/93 & 7/100 & 1/7 & 0/7 \\  
X-X-CHATGPT & 100/100 & 53/100 & 26/100 & 5/100 & 0/100 & 0/0 & 0/0  \\  

\noalign{\vskip 1mm}  
\hline
\end{tabular}
\caption{\label{summ99}
Statistics of the changes in COMET scores with SLoW compared to the baseline of Differ in Round-trip on LLaMa and ChatGPT on the FLORES dataset. Most translation directions have been obviously improved. Details results on BLEU, chrF, and evaluations on DeepSeek-V3 can be found in Table \ref{summemnlp}.
}
\end{table*}
\subsection{Baselines}
We conduct our experiments with both close-sourced and open-sourced LLMs on ChatGPT (GPT-4o-mini), LLaMa-3.1-8B \citep{2024arXiv240721783D} and DeepSeek-V3 671B \citep{2024arXiv241219437D}.  At the time of writing, both of them are popular and widely used English-centric LLMs which are strong in their multilingual translation capacities. Based on these popular LLMs, we compare our proposed method to strong baseline methods:
\begin{itemize}
\setlength\itemsep{0em}
\item \textbf{Vanilla Model} We prompt LLMs to directly translate the input without the assistance of any additional dictionaries.
\item \textbf{Noun Dictionary} Noun words may contain named entities which can be special terminologies which could be particularly hard to translate \citep{ugawa-etal-2018-neural}.
\item \textbf{Adjective Dictionary} Adjective words are another type of word which could be important.
\item \textbf{Verb Dictionary} Verb words are another type of word which could be important.
\item \textbf{Noun and Adjective Dictionary} Combining both noun and adjective dictionaries can be useful as well.
\item \textbf{Noun, Adjective, and Verb Dictionary} Combining noun, adjective, and verb dictionaries can be useful as well.
\item \textbf{Differ in Round-trip} We first use a baseline model without any dictionary to translate the source language into the target language before translating back \citep{sennrich-etal-2016-improving}. The difference between the round-trip translation and the original source sentences is selected as the dictionary.
\item \textbf{Differ in Translation} We first use a baseline without any dictionary to translate the source language into the target language. The difference between the translation and the original target sentences is selected as the dictionary.
\end{itemize}
\subsection{Frequency Estimation}
Since the training sets of LLMs are usually close-sourced, we estimate the word frequency of training data by directly using web resources.\footnote{\url{https://github.com/rspeer/wordfreq}}
\subsection{Prompt Template}
\paragraph{Dictionary Construction} To construct the bilingual dictionary mapping for translation, we prompt ChatGPT \citep{2024arXiv241101141L}:
 
\begin{tcolorbox}
(1) Please provide the translation of the given English sentence into <language>, along with a word-for-word dictionary for each word.\\
(2) The output format must be strictly followed: \\1. Start with `English:' followed by the English sentence. \\2. On the next line, start with `<language>:' followed by the <source> translation. \\3. On the next line, start with `dictionary:' followed by each word in the <language> sentence, annotated with its English meaning in parentheses, separated by spaces.
\\
(3) Now generate translations for the following sentence: 
\\
English: <target>
\\
<language>: <source>
\\ 
dictionary:
\end{tcolorbox}
\paragraph{Translation} We leave the translation prompt in the Appendix due to space reasons.

\input{results.tex}

\section{Conclusions}
LLMs are highly effective in English but underperform in many other languages, especially low-resourced ones. Using dictionary-based methods can improve translation performance, but previous research has not investigated which dictionaries can be more useful to LLMs and they usually add all the dictionaries to the prompt. To this end, we propose a novel method called \textbf{S}elect \textbf{Lo}w-frequency \textbf{W}ords! (\textbf{SLoW}). Given the number of dictionaries to be selected, SLoW selects those with the lowest frequency. We found that such a novel and effective algorithm achieves strong performance, clearly surpassing many strong baselines, including high-frequency dictionaries. Also, general web resources can be used to estimate the frequency instead of the actual training data of the LLMs.
\section*{Limitations}
This paper presents an analysis of 100 languages only. However, there are more than 7,000 languages around the world. The paper can be further extended by including more languages; however, such datasets are lacking. It is quite valuable as the data collection procedure itself is hard, which can significantly contribute to our community.
\par
Second, since we have no access to the training data of the LLMs that we conduct experiments on, it is a pity that we cannot use them to estimate the actual word frequency for experimental purposes.
\section*{Ethical Statement}
We honour and support the ACL ARR Code of Ethics. There is no ethical issue known to us. Well-known and widely used LLMs are used in our work, which is subjected to generating offensive context. Yet, the above-mentioned issues are widely known to exist commonly among LLMs. Any content generated does not reflect the view of the authors.
\bibliography{custom}
\input{appendix}

\end{document}

%% file: results.tex
\section{Results}

\subsection{Main Results}
\paragraph{From-English Translation (EX)} The upper part in Figure \ref{fig:all} visually demonstrates the performance of SLoW on the task of Machine Translation with ADS on the dataset of FLORES compared to strong baselines. The top-left figure shows the performance of LLaMa from English to other languages. The average performance seems to be the lowest among all four figures, which is reasonable. One reason is that this is the translation from English, which is usually lower than translating into English on the English-centric model on average. Another reason is that it is usual for an 8B version LLaMa to be less powerful than close-sourced ChatGPT 4. 
\par
The top-right figure shows the translation performance of ChatGPT from English to other languages. It is obvious that the average performance is better than the from-English direction on LLaMa. This is also reasonable that it is slightly better than the bottom-left figure on translation to English on LLaMa, as to English translation can be considered as generally better than from-English translation on the English-centric model.
\par
Overall, SLoW (purple line) performs clearly better than all the other baselines when translating from English. On LLaMa, it seems that the advantage of SLoW is more clear than on ChatGPT compared to Differ in Round-trip. One postulation is that the performance of LLaMa is generally lower than ChatGPT, so there is more room for improvement for SLoW. Generally speaking, SLoW is clearly useful in improving both LLaMa and ChatGPT on translating from English.
\par
Table \ref{summ99} presents the improvement statistics of SLoW compared to the baseline of Differ in Round-trip for translating from English. SLoW surpasses the baseline. For example, 88 out of 100 language pairs are improved when using SLoW for translating from English for LLaMa. Among those 88 pairs, 22 (25\%) of the pairs are improved for more than 20 COMET scores. In comparison, the number of degradations is apparently lower (12 out of 100 language pairs). When there is a degradation, about half of the language pairs (5/12) give less than 5 points of degradation. These results highlight the usefulness of SLoW.
\par 
\paragraph{Into-English Translation (XE)} The lower part in Figure \ref{fig:all} visually demonstrates the performance of SLoW on the task of Machine Translation with ADS on the dataset of FLORES compared to strong baselines. The bottom-left figure shows the translation performance of LLaMa from English to other languages. The average performance seems to be lower than translating to English on ChatGPT, but higher than translating from English on LLaMa, which is reasonable. One reason is that this is the translation from English, which is usually lower than translating into English on the English-centric model on average. Another reason is that it is usual for an 8B version LLaMa to be less powerful than close-sourced ChatGPT 4. 
\par
\begin{table*}
\centering
\setlength\tabcolsep{5pt}
\setlength\aboverulesep{0pt}\setlength\belowrulesep{0pt}
\setcellgapes{1pt}\makegapedcells
\begin{tabular}{l|l|cc|cc|cc}
\hline
\textbf{PoS}& \textbf{Tag} & \textbf{Per.} & \textbf{Cov.} & \textbf{Per.} & \textbf{Cov.}  & \textbf{Per.} & \textbf{Cov.} \\
\hline
&&\multicolumn{2}{c}{\textit{XE}}&\multicolumn{2}{c}{\textit{EX}}&\multicolumn{2}{c}{\textit{XX}}
\\
\hline
adjective&ADJ&19.28\%&   66.82\% &19.30\%&   58.29\% 
 &19.16\% &   62.88\%\\  
adposition& ADP&2.26\%&    6.10\% & 2.34\%&    6.36\% & 2.51\%&    7.74\%\\  
adverb&ADV& 4.33\%&  41.73\% &  4.13\%&  34.92\% &  4.53\%&  42.93\%\\  
auxiliary& AUX& 0\% &   0\%&  0\% &   0\%&  0\% &   0\%\\  
coordinating conjunction& CCONJ&0.34\% & 4.01\% & 0.15\% & 1.52\%& 0.20\% & 2.20\%  \\  
determiner& DET& 0.45\%&  1.40\% &  0.44\%&  2.41\% &  0.58\%&  3.53\%  \\  
interjection& INTJ& 0\%&  0\% &  0\%&  0\% &  0\%&  0\%  \\  
noun& NOUN& 43.65\%&  49.23\%  &  46.45\%&  45.74\% &  45.61\%&  50.80\% \\  
numeral& NUM& 3.97\%&  53.51\% &  3.35\%&  42.95\%  &  3.34\%&  48.10\%  \\  
particle& PART& 0.11\%&   14.68\% &  0.10\%&   26.32\% & 0.11\%&   32.82\% \\  
pronoun& PRON& 0.68\%& 7.82\% &  0.53\%& 7.03\% &  0.14\%&  90.12\%  \\  
proper noun& PROPN& 0.14\%&  92.15\%  & 0.12\%&  93.83\% &  0.14\%&  90.12\% \\  
punctuation& PUNCT& 0\%& 0\%  &  0\%& 0\% &  0\%& 0\% \\  
subordinating conjunction& SCONJ&0\% &  0\%  &  0\%& 0\% &  0\%& 0\%  \\  
symbol&SYM& 0\% &  0\%&  0\%& 0\% &  0\%& 0\%  \\  
verb& V& 24.06\%& 46.27\% &  22.58\%& 41.85\% &  22.58\%& 47.00\%   \\  
others& X& 0.73\%& 6.71\%  &  0.50\%& 5.85\% & 0.66\%& 8.04\% \\  
 
\hline
\end{tabular}
\caption{\label{pos}
PoS tagger statistics selected by SLoW. Per. represents the percentage of the tag in the whole dictionary prompted, and Cov. represents the coverage, meaning the selected ratio of the selected words compared to the total number of that PoS tag in the dictionary. There are 17 core tags in UPoS: \url{https://universaldependencies.org/u/pos/} UPoS tagger: \url{https://github.com/slavpetrov/universal-pos-tags}, \url{https://www.nltk.org/}.
}
\end{table*}
The bottom-right figure shows the translation performance of ChatGPT into English. It is obvious that the average performance is better than the performance in all the other three figures. This is because translating into English is usually easier than translating from English, and ChatGPT-4 can be usually considered than LLaMa-3.1-8B.
\par
Overall, SLoW (purple line) performs clearly better than all the other baselines for translating from English translation. On LLaMa, it seems that the advantage of SLoW is more clear than on ChatGPT compared to Differ in Round-trip. One postulation is that the performance of LLaMa is generally lower than ChatGPT, so there is more room for improvement for SLoW. Generally speaking, SLoW is clearly useful in improving both LLaMa and ChatGPT on translating from English.
\par
Table \ref{summ99} presents the improvement statistics of SLoW compared to the baseline of Differ in Round-trip for translating into English. It is obvious that SLoW surpasses the baseline. For example, when translating into English on ChatGPT, 92 out of 100 language pairs are improved when using SLoW. Among those 92 pairs, 33 (about 1/3) of the pairs are improved for more than 20 COMET scores. In comparison, the number of degradations is lower (8 out of 100 language pairs). When there is a degradation, about half of the language pairs (5/8) give less than 20 points of degradation. This highlights the usefulness of SLoW.
\par
\paragraph{Non-English-centric Translation (XX)} Figure \ref{fig:xx} visually demonstrates the performance of SLoW on the task of Machine Translation with ADS on the dataset of FLORES compared to strong baselines. The left figure is for translation on LLaMa and the right figure is for translation on ChatGPT. The overall performance on LLaMa is apparently lower than ChatGPT, which is reasonable, as the model size of LLaMa we conduct our experimentation on is obviously smaller than ChatGPT.
\par
Overall, it is clear that SLoW is great at translating in non-English-centric directions, surpassing all the strong baselines. We also note that the baselines might not work well in this scenario, as they can be frequently worse than the Vanilla Baseline, which does not use any additional dictionary. We postulate that this is due to the lack of ability in terms of non-English-centric translation, as ChatGPT is an English-centric model. Still, SLoW is effective and constantly better than the Vanilla Baseline.
\par
Table \ref{summ99} presents the improvement statistics of SLoW compared to the baseline of Differ in Round-trip for translating on non-English-centric directions. It is obvious that SLoW surpasses the baseline. For example, when translating into English on ChatGPT, 100 out of 100 language pairs are improved when using SLoW. Among those 92 pairs, 46 (about half) of the pairs are improved for more than 5 COMET scores. In comparison, the number of degradations is lower, for example, on LLaMa, only 0 out of 100 language pairs are degraded. This highlights the usefulness of SLoW.
\begin{table*}[ht!]
\centering
\setlength\aboverulesep{0pt}\setlength\belowrulesep{0pt}
\setcellgapes{0.5pt}\makegapedcells
    \setlength\tabcolsep{8pt}
    \setlength\extrarowheight{1.5pt}
\begin{tabular}{p{3.7cm}|p{11cm}}

\hline
\textbf{Ground-truth Tsonga} & Muteki wa swifaniso u tleketliwile ku ya le Ronald Reagan UCLA Medical Center, laha a fikeke a fa.   \\
\hline
\textbf{Ground-truth English} & The photographer was transported to Ronald Reagan UCLA Medical Center, where he subsequently died.
\\
\hline
\hline
\textbf{Vanilla Model} & COMET Score: 0.579 \textcolor{red}{\xmark} "Ambulance from Ronald Reagan UCLA Medical Center has arrived." \\
\hline
\hline
\textbf{Differ in Round-trip} & COMET Score: 0.413 \textcolor{red}{\xmark} The ensued entertainment after Ronald Reagan UCLA Medical Centre called the police but attempted him for throad and for traffic stop due to him driving too fast therefore prompting visual imagery for the first time in South Africa. \\
\hline
\hline
\textbf{Differ in Translation} & COMET Score: 0.335 \textcolor{red}{\xmark} The driver of the police vehicle had his ear given according to the witness who is accused of not being prepared to pay a fine. \\
\hline
\hline
\textbf{Noun Dictionary} & COMET Score: 0.806 \textcolor{red}{\xmark} The person in the photographs was taken to the Ronald Reagan UCLA Medical Center, where he died. \\
\hline
\hline
\textbf{Adjective Dictionary} & COMET Score: 0.562 \textcolor{red}{\xmark} Ronald Reagan Center where he died. \\
\hline
\hline
\textbf{Noun and Adjective Dictionary} & COMET Score: 0.763 \textcolor{red}{\xmark} The driver of the vehicle was taken to Ronald Reagan UCLA Medical Center, where he died. \\
\hline
\hline
\textbf{Noun, Adjective, and Verb Dictionary} & COMET Score: 0.864 \textcolor{red}{\xmark} The photographer was taken to the Ronald Reagan UCLA Medical Center, where he died.\\
\hline
\hline
\textbf{High-frequency} & COMET Score: 0.707 \textcolor{red}{\xmark} The man of images was taken to the Ronald Reagan UCLA Medical Center, where he died.\\
\hline
\hline
\textbf{SLoW} & COMET Score: 0.912 \textcolor{green}{\cmark} The photographer was transported to the Ronald Reagan UCLA Medical Center, where he subsequently died.\\
\hline
\end{tabular}
\caption{\label{case2}
A case study on translating from Tsonga To English. \textcolor{red}{\xmark} represents that the generation is not the best among all the models. \textcolor{green}{\cmark} represents that the generation is the best among all the models.
}
\end{table*}

\begin{table*}
\scriptsize
\centering
    \setlength\tabcolsep{1.5pt}
    \setlength\extrarowheight{0.1pt}
\begin{tabular}{l|ccccc|ccccc}
\hline
\noalign{\vskip 1mm}  
\textbf{Direction} & \textbf{\# improved} & \textbf{> 1 point} & \textbf{> 2 points} & \textbf{> 3 points} & \textbf{> 5 points} & \textbf{\# degraded} & \textbf{> 1 point} & \textbf{> 2 points} & \textbf{> 3 points} & \textbf{> 5 points}\\
\noalign{\vskip 1mm}  
\hline
\noalign{\vskip 1mm} 
EX-CHATGPT-BLEU	&72/100&	48/72&	30/72&	20/72&	11/72	&28/100	&9/28	&2/28	&0/28	&0/28\\
EX-CHATGPT-chrF	&69/100	&52/69	&30/69	&20/69	&12/69&	31/100&	13/31&	2/31&	0/31&	0/31\\
XE-CHATGPT-BLEU	&70/100&	50/70&	26/70&	14/70&	11/70&	30/100&	14/30&	8/30&	7/30&	7/30\\
XE-CHATGPT-chrF&	70/100&	41/70&	25/70&	16/70&	13/70&	30/100&	16/30&	9/30&	7/30&	7/30\\
XX-CHATGPT-BLEU&	69/100&	37/69&	19/69&	6/69&	1/69&	31/100&	6/31&	0/31&	0/31&	0/31\\
XX-CHATGPT-chrF&	69/100&	42/69&	19/69&	8/69&	1/69&	31/100&	11/31&	0/31&	0/31&	0/31\\
EX-LLaMa-BLEU&	85/100&	73/85&	59/85&	52/85&	30/85&	15/100&	8/15&	5/15&	5/157&	3/15\\
EX-LLaMa-chrF	&71/100	&50/71	&32/71	&20/71	&13/71&	29/100&	13/29&	5/29&	2/29&	0/29\\
XE-LLaMa-BLEU&	58/100&	37/58&	17/58&	12/58&	10/58&	42/100&	28/42&	11/28&	8/28&	5/28\\
XE-LLaMa-chrF&	64/100&	40/64&	24/64&	18/64&	12/64&	36/100&	24/36&	15/36&	7/36&	6/36\\
XX-LLaMa-BLEU&	84/100&	54/84&	39/84&	28/84&	6/84&	16/100&	6/16&	2/16&	2/16&	0/16\\
XX-LLaMa-chrF&	80/100&	70/80&	50/80&	29/80&	11/80&	20/100&	10/20&	7/20&	5/20&	0/10\\
EX-DEEPSEEKV3-BLEU	&68/100&	47/68&	25/68&	17/68	&12/68&	32/100&	12/32&	2/32&	0/32&	0/32\\
EX-DEEPSEEKV3-chrF&	73/100&	48/73&	31/73&	21/73&	14/73&	27/100&	13/27&	4/27&	1/27&	0/27\\
XE-DEEPSEEKV3-BLEU&	66/100&	44/66&	30/66&	17/66&	13/66&	34/100&	19/34&	9/34&	8/34&	7/34\\
XE-DEEPSEEKV3-chrF&	74/100&	53/74&	35/74&	19/74&	12/74&	26/100&	11/26&	9/26&	9/26&	7/26\\
XX-DEEPSEEKV3-BLEU&	72/100&	38/72&	13/72&	4/72&	0/72&	28/100&	4/28&	0/28&	0/28&	0/28\\
XX-DEEPSEEKV3-chrF&	75/100&	48/75&	25/75&	8/75&	2/75&	25/100&	8/25&	1/25&	0/25&	0/25\\

\noalign{\vskip 1mm}  
\hline
\end{tabular}
\caption{\label{summemnlp}
Statistics of the changes in BLEU and chrF scores with SLoW compared to the baseline of the Noun Dictionary on CHATGPT, LLaMa, and DEEPSEEK-V3. Most translation directions have been obviously improved. In case there is any degradation, the degradation is frequently less than 1 point. 
}
\end{table*}
\subsection{SLoW PoS Tags}
Table \ref{pos} presents the PoS tags of the words selected by SLoW. The dictionary is mainly composed of adjective, noun, and verb words. SLoW surpasses the baseline, which is composed of only these three types of words without considering how frequent they are. In contrast, SLoW selects low-frequency words appropriately, such as numerals and adverbs. However, it could be expensive to run exhaustive experiments on all combinations to be compared with SLoW. Nevertheless, the statistics suggest that SLoW selects a comprehensive dictionary composed of diverse words with different PoS tags, which is effective in improving the translation. This also surpasses the strong baseline with Differ in Round-trip and Differ in Translation.
\par
We present further results on BLEU, chrF evaluations, and results on DeepSeek-V3 in Table \ref{summemnlp}. We also leave case studies in the Table \ref{case2}. They all align with our conclusions. On most language pairs, the performance has been obviously improved. In case there is any degradation, the degradation is frequently less than 1 point. For space reasons, we leave more case studies in our Appendix.
\subsection{SLoW versus Full Dictionary}
\begin{table}
\centering
\setlength\tabcolsep{25pt}
\setlength\aboverulesep{0pt}\setlength\belowrulesep{0pt}
\setcellgapes{3pt}\makegapedcells
\begin{tabular}{l|c}
\hline
\textbf{Direction}& \textbf{Ratio} \\
\hline
into-English&0.553\\  
from-English&0.520\\  
non-English-centric&0.564\\  
\hline
\end{tabular}
\caption{\label{rat}
The dictionary ratio is automatically decided in this paper by aligning with the word numbers in Differ in Round-trip compared to the full dictionary.
}
\end{table}

\begin{table}
\centering
\setlength\tabcolsep{12pt}
\setlength\aboverulesep{0pt}\setlength\belowrulesep{0pt}
\setcellgapes{1pt}\makegapedcells
\begin{tabular}{l|cc}
\hline
\textbf{Direction}& \textbf{SLoW}& \textbf{Full-Dict} \\
\hline
pbt\_Arab&0.803 & 0.483\\
kir\_Cyrl&0.810&0.501\\
gle\_Latn&0.802 & 0.499\\
ory\_Orya&0.839 &  0.518\\
azj\_Latn&0.827 & 0.514\\
\hline
\end{tabular}
\caption{\label{five}
Five XE translation pairs on LLaMa, showing that SLoW obviously surpasses the Full-Dict baseline.
}
\end{table}
While usually adding redundant information to LLMs can degrade performance, removing useful dictionaries can be harmful to translation performance. Table \ref{rat} presents the actual ratio that we have adopted compared to the full dictionary. We also note that under this setting, SLoW can surpass the full dictionary baseline obviously on some language pairs as presented in Table \ref{five}. Yet, for most other cases, the full dictionary is still better, which is however still reasonable and very acceptable as more tokens are cost with LLMs. We also note that there is still a chance for SLoW to surpass the full dictionary baseline better if a different ratio is chosen. Since this is too exhaustive for this paper, we leave the exploration to future work.\footnote{For baselines with more/fewer words than this ratio, random padding or dropping is adopted.}
\begin{table}[t!]
\centering
\begin{tabular}{ccc}
\hline
\noalign{\vskip 1mm}  
\textbf{Direction} & \textbf{High-frequency} & \textbf{SLoW}\\
\noalign{\vskip 1mm}  
\hline
\hline
\noalign{\vskip 1mm}  
XE-ChatGPT & \colorbox{lightgray}{$0$} & \colorbox{cyan}{\textcolor{white}{$\vb*{100}$}}  \\
\noalign{\vskip 1mm} 
EX-ChatGPT & \colorbox{lightgray}{$8$} & \colorbox{cyan}{\textcolor{white}{$\vb*{92}$}}   \\
\noalign{\vskip 1mm} 
XX-ChatGPT & \colorbox{lightgray}{$16$} & \colorbox{cyan}{\textcolor{white}{$\vb*{84}$}}   \\
\noalign{\vskip 1mm} 
XE-LLaMa & \colorbox{lightgray}{$19$} & \colorbox{cyan}{\textcolor{white}{$\vb*{81}$}}  \\
\noalign{\vskip 1mm} 
EX-LLaMa & \colorbox{lightgray}{$1$} & \colorbox{cyan}{\textcolor{white}{$\vb*{99}$}} 
\\
\noalign{\vskip 1mm} 
XX-LLaMa & \colorbox{lightgray}{$13$} & \colorbox{cyan}{\textcolor{white}{$\vb*{87}$}}  \\
\noalign{\vskip 1mm}  
\hline
\hline
\end{tabular}
\caption{\label{high-freq}
The number of winning languages in COMET scores on different language pairs and different models with High-frequency dictionaries and SLoW.
}
\end{table}
\subsection{SLoW versus High-frequency}
In order to further validate our claim that lower-frequency dictionaries are more useful for translation than higher-frequency ones, we perform a comparison between SLoW and those dictionaries with the highest frequency and present the results in Table \ref{high-freq}. When the same number of words and PoS ratios are kept, we see that SLoW is clearly better in high-frequency dictionaries. For example, for translating into English on ChatGPT, SLoW is always better than high-frequency dictionaries. This apparently strengthens the claim of this paper.

%% file: appendix.tex
\pagebreak
\appendix
\onecolumn
\section*{Appendix}
\begin{table*}[htb!]
\centering
\setlength\aboverulesep{0pt}\setlength\belowrulesep{0pt}
\setcellgapes{0pt}\makegapedcells
    \setlength\tabcolsep{7pt}
    \setlength\extrarowheight{4pt}
\begin{tabular}{p{3.7cm}|p{10cm}}

\hline
\textbf{Ground-truth English} & The Iraq Study Group presented its report at 12.00 GMT today.   \\
\hline
\textbf{Ground-truth Standard Malay} & Kumpulan Kajian Iraq telah membentangkan laporan mereka pada pukul 12.00 GMT hari ini.
\\
\hline
\hline
\textbf{Vanilla Model} & COMET Score: 0.944 \textcolor{red}{\xmark} Kumpulan Kajian Iraq melepaskan laporan mereka pada pukul 12.00 GMT hari ini.hobbies \\
\hline
\textbf{Back to English} & The Iraq Study Group released their report at 12:00 GMT today. \\
\hline
\hline
\textbf{Differ in Round-trip} & COMET Score: 0.941 \textcolor{red}{\xmark} Kumpulan Kajian Iraq melaporkan laporan mereka pada pukul 12.00 GMT hari ini. \\
\hline
\textbf{Back to English} & The Iraq Study Group reported their report at 12:00 GMT today. \\
\hline
\hline
\textbf{Differ in Translation} & COMET Score: 0.915 \textcolor{red}{\xmark} Kajian mengenai Iraq telah membentangkan laporan mereka pada jam 1200 pagi ini. \\
\hline
\textbf{Back to English} & The study on Iraq has presented their report at 12:00 this morning. \\
\hline
\hline
\textbf{Noun Dictionary} & COMET Score: 0.949 \textcolor{red}{\xmark} Kumpulan Kajian Iraq memperkenalkan laporan mereka pada jam 12.00 GMT hari ini. \\
\hline
\textbf{Back to English} & The Iraq Study Group introduced their report at 12:00 GMT today. \\
\hline
\hline
\textbf{Adjective Dictionary} & COMET Score: 0.910 \textcolor{red}{\xmark} Kumpulan Kaji Iraq memperkenalkan laporannya pada jam 12.00 GMT pagi ini. \\
\hline
\textbf{Back to English} & The Iraq Study Group introduced its report at 12:00 AM GMT this morning. \\
\hline
\hline
\textbf{Noun and Adjective Dictionary} & COMET Score: 0.946 \textcolor{red}{\xmark} Kumpulan Kajian Iraq memperkenalkan laporannya pada pukul 12.00 GMT hari ini. \\
\hline
\textbf{Back to English} & The Iraq Study Group introduced its report at 12.00 GMT today.\\
\hline
\hline
\textbf{Noun, Adjective, and Verb Dictionary} & COMET Score: 0.894 \textcolor{red}{\xmark} Kumpulan Kaji Iraq mengemukakan laporan mereka pada masa GMT pukul 12 pagi hari ini. \\
\hline
\textbf{Back to English} & The Iraq Study Group submitted their report at 12 AM GMT today.\\
\hline
\hline
\textbf{High-frequency} & COMET Score: 0.940 \textcolor{red}{\xmark} Kumpulan Kajian Iraq memperkenalkan laporannya ini pada jam 12.00 GMT hari ini.\\
\hline
\textbf{Back to English} & The Iraq Study Group introduced its report at 12.00 GMT today.\\
\hline
\hline
\textbf{SLoW} & COMET Score: 0.976 \textcolor{green}{\cmark} Kumpulan Kajian Iraq telah membentangkan laporan mereka pada pukul 12.00 GMT hari ini.\\
\hline
\textbf{Back to English} & The Iraq Study Group has presented their report at 12:00 GMT today.\\
\hline
\end{tabular}
\caption{\label{case1}
A case study on translating from English to Standard Malay. \textcolor{red}{\xmark} represents that the generation is not the best among all the models. \textcolor{green}{\cmark} represents that the generation is the best among all the models.
}
\end{table*}
\twocolumn
\begin{table}[htb]
\centering
\setlength\tabcolsep{8pt}
\setlength\aboverulesep{0pt}\setlength\belowrulesep{0pt}
\setcellgapes{3pt}\makegapedcells
\setlength\extrarowheight{5pt}
\begin{tabular}{c|c}
\hline
\textbf{Language Class} & \textbf{Number}\\
\hline
0&19\\
1&41\\
2&9\\
3&10\\
4&5\\
Total&84\\
\hline
\end{tabular}
\caption{\label{lcc}
A list of language classes of the 100 languages used in our experiments. More than half of the languages used in our study are relatively low-resource according to \citet{joshi-etal-2020-state}.
}
\end{table}

\begin{table}[htb!]
\centering
\setlength\tabcolsep{8pt}
\setlength\aboverulesep{0pt}\setlength\belowrulesep{0pt}
\setcellgapes{3pt}\makegapedcells
\setlength\extrarowheight{1.8pt}
\begin{tabular}{c|c}
\hline
\textbf{Language Pairs} & \textbf{Number}\\
\hline
0 —> 0 &	4\\
0 —> 1&	9\\
0 —> 2&	2\\
0 —> 3&	4\\
0 —> 4&	1\\
1 —> 0	&6\\
1 —> 1	&31\\
1 —> 2	&7\\
1 —> 3	&4\\
1 —> 4	&1\\
2 —> 0	&3\\
2 —> 1&	8\\
2 —> 2	&0\\
2 —> 3	&1\\
2 —> 4	&2\\
3 —> 0	&2\\
3 —> 1	&7\\
3 —> 2	&2\\
3 —> 3	&0\\
3 —> 4	&0\\
4 —> 0	&1\\
4 —> 1	&3\\
4 —> 2	&0\\
4 —> 3	&2\\
4 —> 4	&0\\
Total	&100\\

\hline
\end{tabular}
\caption{\label{lcc2}
A list of language pair classes of the XX translation experiments. More than half of the languages used in our study are relatively low-resource according to \citet{joshi-etal-2020-state}.
}
\end{table}
\par
We use the following prompt for translation:
\begin{tcolorbox}
Translate the following sentence from \{source\_language\} to \{target\_language\}.\\
\{origin\_sentence\}\\
Use the provided dictionary to clarify or improve the translation of any misaligned words.\\
- Here are some dictionaries that you need to focus on:\\
\{dict\}\\
Note: Finally, only respond to me with the final \{target\_language\} translation. Your output format is as follows:\\
The refined translation is:
\end{tcolorbox}
The dictionary size for the constructed dictionary is: EX: 1581.62, XE: 1539.72, XX: 1636.59, averaged from all languages in our experiments. For each prompt, the number of dictionaries to be used in all models is aligned with the baseline Differ in Round-trip throughout experiments.

%% file: custom.bib
@ARTICLE{2024arXiv241219437D,
       author = {{DeepSeek-AI} and {Liu}, Aixin and {Feng}, Bei and {Xue}, Bing and {Wang}, Bingxuan and {Wu}, Bochao and {Lu}, Chengda and {Zhao}, Chenggang and {Deng}, Chengqi and {Zhang}, Chenyu and {Ruan}, Chong and {Dai}, Damai and {Guo}, Daya and {Yang}, Dejian and {Chen}, Deli and {Ji}, Dongjie and {Li}, Erhang and {Lin}, Fangyun and {Dai}, Fucong and {Luo}, Fuli and {Hao}, Guangbo and {Chen}, Guanting and {Li}, Guowei and {Zhang}, H. and {Bao}, Han and {Xu}, Hanwei and {Wang}, Haocheng and {Zhang}, Haowei and {Ding}, Honghui and {Xin}, Huajian and {Gao}, Huazuo and {Li}, Hui and {Qu}, Hui and {Cai}, J.~L. and {Liang}, Jian and {Guo}, Jianzhong and {Ni}, Jiaqi and {Li}, Jiashi and {Wang}, Jiawei and {Chen}, Jin and {Chen}, Jingchang and {Yuan}, Jingyang and {Qiu}, Junjie and {Li}, Junlong and {Song}, Junxiao and {Dong}, Kai and {Hu}, Kai and {Gao}, Kaige and {Guan}, Kang and {Huang}, Kexin and {Yu}, Kuai and {Wang}, Lean and {Zhang}, Lecong and {Xu}, Lei and {Xia}, Leyi and {Zhao}, Liang and {Wang}, Litong and {Zhang}, Liyue and {Li}, Meng and {Wang}, Miaojun and {Zhang}, Mingchuan and {Zhang}, Minghua and {Tang}, Minghui and {Li}, Mingming and {Tian}, Ning and {Huang}, Panpan and {Wang}, Peiyi and {Zhang}, Peng and {Wang}, Qiancheng and {Zhu}, Qihao and {Chen}, Qinyu and {Du}, Qiushi and {Chen}, R.~J. and {Jin}, R.~L. and {Ge}, Ruiqi and {Zhang}, Ruisong and {Pan}, Ruizhe and {Wang}, Runji and {Xu}, Runxin and {Zhang}, Ruoyu and {Chen}, Ruyi and {Li}, S.~S. and {Lu}, Shanghao and {Zhou}, Shangyan and {Chen}, Shanhuang and {Wu}, Shaoqing and {Ye}, Shengfeng and {Ye}, Shengfeng and {Ma}, Shirong and {Wang}, Shiyu and {Zhou}, Shuang and {Yu}, Shuiping and {Zhou}, Shunfeng and {Pan}, Shuting and {Wang}, T. and {Yun}, Tao and {Pei}, Tian and {Sun}, Tianyu and {Xiao}, W.~L. and {Zeng}, Wangding and {Zhao}, Wanjia and {An}, Wei and {Liu}, Wen and {Liang}, Wenfeng and {Gao}, Wenjun and {Yu}, Wenqin and {Zhang}, Wentao and {Li}, X.~Q. and {Jin}, Xiangyue and {Wang}, Xianzu and {Bi}, Xiao and {Liu}, Xiaodong and {Wang}, Xiaohan and {Shen}, Xiaojin and {Chen}, Xiaokang and {Zhang}, Xiaokang and {Chen}, Xiaosha and {Nie}, Xiaotao and {Sun}, Xiaowen and {Wang}, Xiaoxiang and {Cheng}, Xin and {Liu}, Xin and {Xie}, Xin and {Liu}, Xingchao and {Yu}, Xingkai and {Song}, Xinnan and {Shan}, Xinxia and {Zhou}, Xinyi and {Yang}, Xinyu and {Li}, Xinyuan and {Su}, Xuecheng and {Lin}, Xuheng and {Li}, Y.~K. and {Wang}, Y.~Q. and {Wei}, Y.~X. and {Zhu}, Y.~X. and {Zhang}, Yang and {Xu}, Yanhong and {Xu}, Yanhong and {Huang}, Yanping and {Li}, Yao and {Zhao}, Yao and {Sun}, Yaofeng and {Li}, Yaohui and {Wang}, Yaohui and {Yu}, Yi and {Zheng}, Yi and {Zhang}, Yichao and {Shi}, Yifan and {Xiong}, Yiliang and {He}, Ying and {Tang}, Ying and {Piao}, Yishi and {Wang}, Yisong and {Tan}, Yixuan and {Ma}, Yiyang and {Liu}, Yiyuan and {Guo}, Yongqiang and {Wu}, Yu and {Ou}, Yuan and {Zhu}, Yuchen and {Wang}, Yuduan and {Gong}, Yue and {Zou}, Yuheng and {He}, Yujia and {Zha}, Yukun and {Xiong}, Yunfan and {Ma}, Yunxian and {Yan}, Yuting and {Luo}, Yuxiang and {You}, Yuxiang and {Liu}, Yuxuan and {Zhou}, Yuyang and {Wu}, Z.~F. and {Ren}, Z.~Z. and {Ren}, Zehui and {Sha}, Zhangli and {Fu}, Zhe and {Xu}, Zhean and {Huang}, Zhen and {Zhang}, Zhen and {Xie}, Zhenda and {Zhang}, Zhengyan and {Hao}, Zhewen and {Gou}, Zhibin and {Ma}, Zhicheng and {Yan}, Zhigang and {Shao}, Zhihong and {Xu}, Zhipeng and {Wu}, Zhiyu and {Zhang}, Zhongyu and {Li}, Zhuoshu and {Gu}, Zihui and {Zhu}, Zijia and {Liu}, Zijun and {Li}, Zilin and {Xie}, Ziwei and {Song}, Ziyang and {Gao}, Ziyi and {Pan}, Zizheng},
        title = "{DeepSeek-V3 Technical Report}",
      journal = {arXiv e-prints},
         year = 2024,
        month = dec,
          eid = {arXiv:2412.19437},
        pages = {arXiv:2412.19437},
          doi = {10.48550/arXiv.2412.19437},
archivePrefix = {arXiv},
       eprint = {2412.19437},
 primaryClass = {cs.CL},
       adsurl = {https://ui.adsabs.harvard.edu/abs/2024arXiv241219437D}
}

@inproceedings{sennrich-etal-2016-improving,
    title = "Improving Neural Machine Translation Models with Monolingual Data",
    author = "Sennrich, Rico  and
      Haddow, Barry  and
      Birch, Alexandra",
    editor = "Erk, Katrin  and
      Smith, Noah A.",
    booktitle = "Proceedings of the 54th Annual Meeting of the Association for Computational Linguistics (Volume 1: Long Papers)",
    month = aug,
    year = "2016",
    address = "Berlin, Germany",
    publisher = "Association for Computational Linguistics",
    url = "https://aclanthology.org/P16-1009",
    doi = "10.18653/v1/P16-1009",
    pages = "86--96",
}

@inproceedings{joshi-etal-2020-state,
    title = "The State and Fate of Linguistic Diversity and Inclusion in the {NLP} World",
    author = "Joshi, Pratik  and
      Santy, Sebastin  and
      Budhiraja, Amar  and
      Bali, Kalika  and
      Choudhury, Monojit",
    editor = "Jurafsky, Dan  and
      Chai, Joyce  and
      Schluter, Natalie  and
      Tetreault, Joel",
    booktitle = "Proceedings of the 58th Annual Meeting of the Association for Computational Linguistics",
    month = jul,
    year = "2020",
    address = "Online",
    publisher = "Association for Computational Linguistics",
    url = "https://aclanthology.org/2020.acl-main.560/",
    doi = "10.18653/v1/2020.acl-main.560",
    pages = "6282--6293"
}

@inproceedings{ugawa-etal-2018-neural,
    title = "Neural Machine Translation Incorporating Named Entity",
    author = "Ugawa, Arata  and
      Tamura, Akihiro  and
      Ninomiya, Takashi  and
      Takamura, Hiroya  and
      Okumura, Manabu",
    editor = "Bender, Emily M.  and
      Derczynski, Leon  and
      Isabelle, Pierre",
    booktitle = "Proceedings of the 27th International Conference on Computational Linguistics",
    month = aug,
    year = "2018",
    address = "Santa Fe, New Mexico, USA",
    publisher = "Association for Computational Linguistics",
    url = "https://aclanthology.org/C18-1274",
    pages = "3240--3250",
}

@inproceedings{rei-etal-2020-comet,
    title = "{COMET}: A Neural Framework for {MT} Evaluation",
    author = "Rei, Ricardo  and
      Stewart, Craig  and
      Farinha, Ana C  and
      Lavie, Alon",
    booktitle = "Proceedings of the 2020 Conference on Empirical Methods in Natural Language Processing (EMNLP)",
    month = nov,
    year = "2020",
    address = "Online",
    publisher = "Association for Computational Linguistics",
    url = "https://aclanthology.org/2020.emnlp-main.213",
    doi = "10.18653/v1/2020.emnlp-main.213",
    pages = "2685--2702",
}

@inproceedings{seq2seq,
author = {Sutskever, Ilya and Vinyals, Oriol and Le, Quoc V.},
title = {Sequence to Sequence Learning with Neural Networks},
year = {2014},
publisher = {MIT Press},
address = {Cambridge, MA, USA},
booktitle = {Proceedings of the 27th International Conference on Neural Information Processing Systems - Volume 2},
pages = {3104–3112},
numpages = {9},
location = {Montreal, Canada},
series = {NIPS'14}
}

@inproceedings{10.5555/3618408.3619699,
author = {Shi, Freda and Chen, Xinyun and Misra, Kanishka and Scales, Nathan and Dohan, David and Chi, Ed and Sch\"{a}rli, Nathanael and Zhou, Denny},
title = {Large language models can be easily distracted by irrelevant context},
year = {2023},
publisher = {JMLR.org},
booktitle = {Proceedings of the 40th International Conference on Machine Learning},
articleno = {1291},
numpages = {18},
location = {Honolulu, Hawaii, USA},
series = {ICML'23}
}

@ARTICLE{2024arXiv241101141L,
       author = {{Lu}, Hongyuan and {Li}, Zixuan and {Lam}, Wai},
        title = "{Dictionary Insertion Prompting for Multilingual Reasoning on Multilingual Large Language Models}",
      journal = {arXiv e-prints},
         year = 2024,
        month = nov,
          eid = {arXiv:2411.01141},
        pages = {arXiv:2411.01141},
          doi = {10.48550/arXiv.2411.01141},
archivePrefix = {arXiv},
       eprint = {2411.01141},
 primaryClass = {cs.CL},
       adsurl = {https://ui.adsabs.harvard.edu/abs/2024arXiv241101141L}
}

@inproceedings{lu-etal-2024-chain,
    title = "Chain-of-Dictionary Prompting Elicits Translation in Large Language Models",
    author = "Lu, Hongyuan  and
      Yang, Haoran  and
      Huang, Haoyang  and
      Zhang, Dongdong  and
      Lam, Wai  and
      Wei, Furu",
    editor = "Al-Onaizan, Yaser  and
      Bansal, Mohit  and
      Chen, Yun-Nung",
    booktitle = "Proceedings of the 2024 Conference on Empirical Methods in Natural Language Processing",
    month = nov,
    year = "2024",
    address = "Miami, Florida, USA",
    publisher = "Association for Computational Linguistics",
    url = "https://aclanthology.org/2024.emnlp-main.55",
    doi = "10.18653/v1/2024.emnlp-main.55",
    pages = "958--976",
}

@inproceedings{popovic-2015-chrf,
    title = "chr{F}: character n-gram {F}-score for automatic {MT} evaluation",
    author = "Popovi{\'c}, Maja",
    booktitle = "Proceedings of the Tenth Workshop on Statistical Machine Translation",
    month = sep,
    year = "2015",
    address = "Lisbon, Portugal",
    publisher = "Association for Computational Linguistics",
    url = "https://aclanthology.org/W15-3049",
    doi = "10.18653/v1/W15-3049",
    pages = "392--395",
}

@inproceedings{papineni-etal-2002-bleu,
    title = "{B}leu: a Method for Automatic Evaluation of Machine Translation",
    author = "Papineni, Kishore  and
      Roukos, Salim  and
      Ward, Todd  and
      Zhu, Wei-Jing",
    editor = "Isabelle, Pierre  and
      Charniak, Eugene  and
      Lin, Dekang",
    booktitle = "Proceedings of the 40th Annual Meeting of the Association for Computational Linguistics",
    month = jul,
    year = "2002",
    address = "Philadelphia, Pennsylvania, USA",
    publisher = "Association for Computational Linguistics",
    url = "https://aclanthology.org/P02-1040",
    doi = "10.3115/1073083.1073135",
    pages = "311--318",
}

@ARTICLE{2024arXiv240721783D,
       author = {{Dubey}, Abhimanyu and {Jauhri}, Abhinav and {Pandey}, Abhinav and {Kadian}, Abhishek and {Al-Dahle}, Ahmad and {Letman}, Aiesha and {Mathur}, Akhil and {Schelten}, Alan and {Yang}, Amy and {Fan}, Angela and {Goyal}, Anirudh and {Hartshorn}, Anthony and {Yang}, Aobo and {Mitra}, Archi and {Sravankumar}, Archie and {Korenev}, Artem and {Hinsvark}, Arthur and {Rao}, Arun and {Zhang}, Aston and {Rodriguez}, Aurelien and {Gregerson}, Austen and {Spataru}, Ava and {Roziere}, Baptiste and {Biron}, Bethany and {Tang}, Binh and {Chern}, Bobbie and {Caucheteux}, Charlotte and {Nayak}, Chaya and {Bi}, Chloe and {Marra}, Chris and {McConnell}, Chris and {Keller}, Christian and {Touret}, Christophe and {Wu}, Chunyang and {Wong}, Corinne and {Canton Ferrer}, Cristian and {Nikolaidis}, Cyrus and {Allonsius}, Damien and {Song}, Daniel and {Pintz}, Danielle and {Livshits}, Danny and {Esiobu}, David and {Choudhary}, Dhruv and {Mahajan}, Dhruv and {Garcia-Olano}, Diego and {Perino}, Diego and {Hupkes}, Dieuwke and {Lakomkin}, Egor and {AlBadawy}, Ehab and {Lobanova}, Elina and {Dinan}, Emily and {Smith}, Eric Michael and {Radenovic}, Filip and {Zhang}, Frank and {Synnaeve}, Gabriel and {Lee}, Gabrielle and {Anderson}, Georgia Lewis and {Nail}, Graeme and {Mialon}, Gregoire and {Pang}, Guan and {Cucurell}, Guillem and {Nguyen}, Hailey and {Korevaar}, Hannah and {Xu}, Hu and {Touvron}, Hugo and {Zarov}, Iliyan and {Arrieta Ibarra}, Imanol and {Kloumann}, Isabel and {Misra}, Ishan and {Evtimov}, Ivan and {Copet}, Jade and {Lee}, Jaewon and {Geffert}, Jan and {Vranes}, Jana and {Park}, Jason and {Mahadeokar}, Jay and {Shah}, Jeet and {van der Linde}, Jelmer and {Billock}, Jennifer and {Hong}, Jenny and {Lee}, Jenya and {Fu}, Jeremy and {Chi}, Jianfeng and {Huang}, Jianyu and {Liu}, Jiawen and {Wang}, Jie and {Yu}, Jiecao and {Bitton}, Joanna and {Spisak}, Joe and {Park}, Jongsoo and {Rocca}, Joseph and {Johnstun}, Joshua and {Saxe}, Joshua and {Jia}, Junteng and {Vasuden Alwala}, Kalyan and {Upasani}, Kartikeya and {Plawiak}, Kate and {Li}, Ke and {Heafield}, Kenneth and {Stone}, Kevin and {El-Arini}, Khalid and {Iyer}, Krithika and {Malik}, Kshitiz and {Chiu}, Kuenley and {Bhalla}, Kunal and {Rantala-Yeary}, Lauren and {van der Maaten}, Laurens and {Chen}, Lawrence and {Tan}, Liang and {Jenkins}, Liz and {Martin}, Louis and {Madaan}, Lovish and {Malo}, Lubo and {Blecher}, Lukas and {Landzaat}, Lukas and {de Oliveira}, Luke and {Muzzi}, Madeline and {Pasupuleti}, Mahesh and {Singh}, Mannat and {Paluri}, Manohar and {Kardas}, Marcin and {Oldham}, Mathew and {Rita}, Mathieu and {Pavlova}, Maya and {Kambadur}, Melanie and {Lewis}, Mike and {Si}, Min and {Singh}, Mitesh Kumar and {Hassan}, Mona and {Goyal}, Naman and {Torabi}, Narjes and {Bashlykov}, Nikolay and {Bogoychev}, Nikolay and {Chatterji}, Niladri and {Duchenne}, Olivier and {{\c{C}}elebi}, Onur and {Alrassy}, Patrick and {Zhang}, Pengchuan and {Li}, Pengwei and {Vasic}, Petar and {Weng}, Peter and {Bhargava}, Prajjwal and {Dubal}, Pratik and {Krishnan}, Praveen and {Singh Koura}, Punit and {Xu}, Puxin and {He}, Qing and {Dong}, Qingxiao and {Srinivasan}, Ragavan and {Ganapathy}, Raj and {Calderer}, Ramon and {Silveira Cabral}, Ricardo and {Stojnic}, Robert and {Raileanu}, Roberta and {Girdhar}, Rohit and {Patel}, Rohit and {Sauvestre}, Romain and {Polidoro}, Ronnie and {Sumbaly}, Roshan and {Taylor}, Ross and {Silva}, Ruan and {Hou}, Rui and {Wang}, Rui and {Hosseini}, Saghar and {Chennabasappa}, Sahana and {Singh}, Sanjay and {Bell}, Sean and {Kim}, Seohyun Sonia and {Edunov}, Sergey and {Nie}, Shaoliang and {Narang}, Sharan and {Raparthy}, Sharath and {Shen}, Sheng and {Wan}, Shengye and {Bhosale}, Shruti and {Zhang}, Shun and {Vandenhende}, Simon and {Batra}, Soumya and {Whitman}, Spencer and {Sootla}, Sten and {Collot}, Stephane and {Gururangan}, Suchin and {Borodinsky}, Sydney and {Herman}, Tamar and {Fowler}, Tara and {Sheasha}, Tarek and {Georgiou}, Thomas and {Scialom}, Thomas and {Speckbacher}, Tobias and {Mihaylov}, Todor and {Xiao}, Tong and {Karn}, Ujjwal and {Goswami}, Vedanuj and {Gupta}, Vibhor and {Ramanathan}, Vignesh and {Kerkez}, Viktor and {Gonguet}, Vincent and {Do}, Virginie and {Vogeti}, Vish and {Petrovic}, Vladan and {Chu}, Weiwei and {Xiong}, Wenhan and {Fu}, Wenyin and {Meers}, Whitney and {Martinet}, Xavier and {Wang}, Xiaodong and {Tan}, Xiaoqing Ellen and {Xie}, Xinfeng and {Jia}, Xuchao and {Wang}, Xuewei and {Goldschlag}, Yaelle and {Gaur}, Yashesh and {Babaei}, Yasmine and {Wen}, Yi and {Song}, Yiwen and {Zhang}, Yuchen and {Li}, Yue and {Mao}, Yuning and {Delpierre Coudert}, Zacharie and {Yan}, Zheng and {Chen}, Zhengxing and {Papakipos}, Zoe and {Singh}, Aaditya and {Grattafiori}, Aaron and {Jain}, Abha and {Kelsey}, Adam and {Shajnfeld}, Adam and {Gangidi}, Adithya and {Victoria}, Adolfo and {Goldstand}, Ahuva and {Menon}, Ajay and {Sharma}, Ajay and {Boesenberg}, Alex and {Vaughan}, Alex and {Baevski}, Alexei and {Feinstein}, Allie and {Kallet}, Amanda and {Sangani}, Amit and {Yunus}, Anam and {Lupu}, Andrei and {Alvarado}, Andres and {Caples}, Andrew and {Gu}, Andrew and {Ho}, Andrew and {Poulton}, Andrew and {Ryan}, Andrew and {Ramchandani}, Ankit and {Franco}, Annie and {Saraf}, Aparajita and {Chowdhury}, Arkabandhu and {Gabriel}, Ashley and {Bharambe}, Ashwin and {Eisenman}, Assaf and {Yazdan}, Azadeh and {James}, Beau and {Maurer}, Ben and {Leonhardi}, Benjamin and {Huang}, Bernie and {Loyd}, Beth and {De Paola}, Beto and {Paranjape}, Bhargavi and {Liu}, Bing and {Wu}, Bo and {Ni}, Boyu and {Hancock}, Braden and {Wasti}, Bram and {Spence}, Brandon and {Stojkovic}, Brani and {Gamido}, Brian and {Montalvo}, Britt and {Parker}, Carl and {Burton}, Carly and {Mejia}, Catalina and {Wang}, Changhan and {Kim}, Changkyu and {Zhou}, Chao and {Hu}, Chester and {Chu}, Ching-Hsiang and {Cai}, Chris and {Tindal}, Chris and {Feichtenhofer}, Christoph and {Civin}, Damon and {Beaty}, Dana and {Kreymer}, Daniel and {Li}, Daniel and {Wyatt}, Danny and {Adkins}, David and {Xu}, David and {Testuggine}, Davide and {David}, Delia and {Parikh}, Devi and {Liskovich}, Diana and {Foss}, Didem and {Wang}, Dingkang and {Le}, Duc and {Holland}, Dustin and {Dowling}, Edward and {Jamil}, Eissa and {Montgomery}, Elaine and {Presani}, Eleonora and {Hahn}, Emily and {Wood}, Emily and {Brinkman}, Erik and {Arcaute}, Esteban and {Dunbar}, Evan and {Smothers}, Evan and {Sun}, Fei and {Kreuk}, Felix and {Tian}, Feng and {Ozgenel}, Firat and {Caggioni}, Francesco and {Guzm{\'a}n}, Francisco and {Kanayet}, Frank and {Seide}, Frank and {Medina Florez}, Gabriela and {Schwarz}, Gabriella and {Badeer}, Gada and {Swee}, Georgia and {Halpern}, Gil and {Thattai}, Govind and {Herman}, Grant and {Sizov}, Grigory and {Guangyi} and {Zhang} and {Lakshminarayanan}, Guna and {Shojanazeri}, Hamid and {Zou}, Han and {Wang}, Hannah and {Zha}, Hanwen and {Habeeb}, Haroun and {Rudolph}, Harrison and {Suk}, Helen and {Aspegren}, Henry and {Goldman}, Hunter and {Damlaj}, Ibrahim and {Molybog}, Igor and {Tufanov}, Igor and {Veliche}, Irina-Elena and {Gat}, Itai and {Weissman}, Jake and {Geboski}, James and {Kohli}, James and {Asher}, Japhet and {Gaya}, Jean-Baptiste and {Marcus}, Jeff and {Tang}, Jeff and {Chan}, Jennifer and {Zhen}, Jenny and {Reizenstein}, Jeremy and {Teboul}, Jeremy and {Zhong}, Jessica and {Jin}, Jian and {Yang}, Jingyi and {Cummings}, Joe and {Carvill}, Jon and {Shepard}, Jon and {McPhie}, Jonathan and {Torres}, Jonathan and {Ginsburg}, Josh and {Wang}, Junjie and {Wu}, Kai and {Hou U}, Kam and {Saxena}, Karan and {Prasad}, Karthik and {Khandelwal}, Kartikay and {Zand}, Katayoun and {Matosich}, Kathy and {Veeraraghavan}, Kaushik and {Michelena}, Kelly and {Li}, Keqian and {Huang}, Kun and {Chawla}, Kunal and {Lakhotia}, Kushal and {Huang}, Kyle and {Chen}, Lailin and {Garg}, Lakshya and {A}, Lavender and {Silva}, Leandro and {Bell}, Lee and {Zhang}, Lei and {Guo}, Liangpeng and {Yu}, Licheng and {Moshkovich}, Liron and {Wehrstedt}, Luca and {Khabsa}, Madian and {Avalani}, Manav and {Bhatt}, Manish and {Tsimpoukelli}, Maria and {Mankus}, Martynas and {Hasson}, Matan and {Lennie}, Matthew and {Reso}, Matthias and {Groshev}, Maxim and {Naumov}, Maxim and {Lathi}, Maya and {Keneally}, Meghan and {Seltzer}, Michael L. and {Valko}, Michal and {Restrepo}, Michelle and {Patel}, Mihir and {Vyatskov}, Mik and {Samvelyan}, Mikayel and {Clark}, Mike and {Macey}, Mike and {Wang}, Mike and {Jubert Hermoso}, Miquel and {Metanat}, Mo and {Rastegari}, Mohammad and {Bansal}, Munish and {Santhanam}, Nandhini and {Parks}, Natascha and {White}, Natasha and {Bawa}, Navyata and {Singhal}, Nayan and {Egebo}, Nick and {Usunier}, Nicolas and {Pavlovich Laptev}, Nikolay and {Dong}, Ning and {Zhang}, Ning and {Cheng}, Norman and {Chernoguz}, Oleg and {Hart}, Olivia and {Salpekar}, Omkar and {Kalinli}, Ozlem and {Kent}, Parkin and {Parekh}, Parth and {Saab}, Paul and {Balaji}, Pavan and {Rittner}, Pedro and {Bontrager}, Philip and {Roux}, Pierre and {Dollar}, Piotr and {Zvyagina}, Polina and {Ratanchandani}, Prashant and {Yuvraj}, Pritish and {Liang}, Qian and {Alao}, Rachad and {Rodriguez}, Rachel and {Ayub}, Rafi and {Murthy}, Raghotham and {Nayani}, Raghu and {Mitra}, Rahul and {Li}, Raymond and {Hogan}, Rebekkah and {Battey}, Robin and {Wang}, Rocky and {Maheswari}, Rohan and {Howes}, Russ and {Rinott}, Ruty and {Jayesh Bondu}, Sai and {Datta}, Samyak and {Chugh}, Sara and {Hunt}, Sara and {Dhillon}, Sargun and {Sidorov}, Sasha and {Pan}, Satadru and {Verma}, Saurabh and {Yamamoto}, Seiji and {Ramaswamy}, Sharadh and {Lindsay}, Shaun and {Lindsay}, Shaun and {Feng}, Sheng and {Lin}, Shenghao and {Zha}, Shengxin Cindy and {Shankar}, Shiva and {Zhang}, Shuqiang and {Zhang}, Shuqiang and {Wang}, Sinong and {Agarwal}, Sneha and {Sajuyigbe}, Soji and {Chintala}, Soumith and {Max}, Stephanie and {Chen}, Stephen and {Kehoe}, Steve and {Satterfield}, Steve and {Govindaprasad}, Sudarshan and {Gupta}, Sumit and {Cho}, Sungmin and {Virk}, Sunny and {Subramanian}, Suraj and {Choudhury}, Sy and {Goldman}, Sydney and {Remez}, Tal and {Glaser}, Tamar and {Best}, Tamara and {Kohler}, Thilo and {Robinson}, Thomas and {Li}, Tianhe and {Zhang}, Tianjun and {Matthews}, Tim and {Chou}, Timothy and {Shaked}, Tzook and {Vontimitta}, Varun and {Ajayi}, Victoria and {Montanez}, Victoria and {Mohan}, Vijai and {Satish Kumar}, Vinay and {Mangla}, Vishal and {Albiero}, V{\'\i}tor and {Ionescu}, Vlad and {Poenaru}, Vlad and {Tiberiu Mihailescu}, Vlad and {Ivanov}, Vladimir and {Li}, Wei and {Wang}, Wenchen and {Jiang}, Wenwen and {Bouaziz}, Wes and {Constable}, Will and {Tang}, Xiaocheng and {Wang}, Xiaofang and {Wu}, Xiaojian and {Wang}, Xiaolan and {Xia}, Xide and {Wu}, Xilun and {Gao}, Xinbo and {Chen}, Yanjun and {Hu}, Ye and {Jia}, Ye and {Qi}, Ye and {Li}, Yenda and {Zhang}, Yilin and {Zhang}, Ying and {Adi}, Yossi and {Nam}, Youngjin and {Yu} and {Wang} and {Hao}, Yuchen and {Qian}, Yundi and {He}, Yuzi and {Rait}, Zach and {DeVito}, Zachary and {Rosnbrick}, Zef and {Wen}, Zhaoduo and {Yang}, Zhenyu and {Zhao}, Zhiwei},
        title = "{The Llama 3 Herd of Models}",
      journal = {arXiv e-prints},
         year = 2024,
        month = jul,
          eid = {arXiv:2407.21783},
        pages = {arXiv:2407.21783},
          doi = {10.48550/arXiv.2407.21783},
archivePrefix = {arXiv},
       eprint = {2407.21783},
 primaryClass = {cs.AI},
       adsurl = {https://ui.adsabs.harvard.edu/abs/2024arXiv240721783D}
}

@inproceedings{10.1145/3377713.3377801,
author = {H\"{a}m\"{a}l\"{a}inen, Mika and Alnajjar, Khalid},
title = {A Template Based Approach for Training NMT for Low-Resource Uralic Languages - A Pilot with Finnish},
year = {2020},
isbn = {9781450372619},
publisher = {Association for Computing Machinery},
address = {New York, NY, USA},
url = {https://doi.org/10.1145/3377713.3377801},
doi = {10.1145/3377713.3377801},
booktitle = {Proceedings of the 2019 2nd International Conference on Algorithms, Computing and Artificial Intelligence},
pages = {520–525},
numpages = {6},
location = {Sanya, China},
series = {ACAI '19}
}

@inproceedings{arthur-etal-2016-incorporating,
    title = "Incorporating Discrete Translation Lexicons into Neural Machine Translation",
    author = "Arthur, Philip  and
      Neubig, Graham  and
      Nakamura, Satoshi",
    booktitle = "Proceedings of the 2016 Conference on Empirical Methods in Natural Language Processing",
    month = nov,
    year = "2016",
    address = "Austin, Texas",
    publisher = "Association for Computational Linguistics",
    url = "https://aclanthology.org/D16-1162",
    doi = "10.18653/v1/D16-1162",
    pages = "1557--1567",
}

@ARTICLE{2016arXiv161007272Z,
       author = {{Zhang}, Jiajun and {Zong}, Chengqing},
        title = "{Bridging Neural Machine Translation and Bilingual Dictionaries}",
      journal = {arXiv e-prints},
         year = 2016,
        month = oct,
          eid = {arXiv:1610.07272},
        pages = {arXiv:1610.07272},
          doi = {10.48550/arXiv.1610.07272},
archivePrefix = {arXiv},
       eprint = {1610.07272},
 primaryClass = {cs.CL},
       adsurl = {https://ui.adsabs.harvard.edu/abs/2016arXiv161007272Z}
}

@inproceedings{10.5555/3491440.3491936,
author = {Chen, Guanhua and Chen, Yun and Wang, Yong and Li, Victor O. K.},
title = {Lexical-Constraint-Aware Neural Machine Translation via Data Augmentation},
year = {2021},
isbn = {9780999241165},
booktitle = {Proceedings of the Twenty-Ninth International Joint Conference on Artificial Intelligence},
articleno = {496},
numpages = {7},
location = {Yokohama, Yokohama, Japan},
series = {IJCAI'20}
}

@inproceedings{dinu-etal-2019-training,
    title = "Training Neural Machine Translation to Apply Terminology Constraints",
    author = "Dinu, Georgiana  and
      Mathur, Prashant  and
      Federico, Marcello  and
      Al-Onaizan, Yaser",
    booktitle = "Proceedings of the 57th Annual Meeting of the Association for Computational Linguistics",
    month = jul,
    year = "2019",
    address = "Florence, Italy",
    publisher = "Association for Computational Linguistics",
    url = "https://aclanthology.org/P19-1294",
    doi = "10.18653/v1/P19-1294",
    pages = "3063--3068",
}

@inproceedings{2019arXiv190409107S,
    title = "Code-Switching for Enhancing {NMT} with Pre-Specified Translation",
    author = "Song, Kai  and
      Zhang, Yue  and
      Yu, Heng  and
      Luo, Weihua  and
      Wang, Kun  and
      Zhang, Min",
    booktitle = "Proceedings of the 2019 Conference of the North {A}merican Chapter of the Association for Computational Linguistics: Human Language Technologies, Volume 1 (Long and Short Papers)",
    month = jun,
    year = "2019",
    address = "Minneapolis, Minnesota",
    publisher = "ACL",
    url = "https://aclanthology.org/N19-1044",
    doi = "10.18653/v1/N19-1044",
    pages = "449--459",
}

@inproceedings{post-vilar-2018-fast,
    title = "Fast Lexically Constrained Decoding with Dynamic Beam Allocation for Neural Machine Translation",
    author = "Post, Matt  and
      Vilar, David",
    booktitle = "Proceedings of the 2018 Conference of the North {A}merican Chapter of the Association for Computational Linguistics: Human Language Technologies, Volume 1 (Long Papers)",
    month = jun,
    year = "2018",
    address = "New Orleans, Louisiana",
    publisher = "Association for Computational Linguistics",
    url = "https://aclanthology.org/N18-1119",
    doi = "10.18653/v1/N18-1119",
    pages = "1314--1324",
}

@inproceedings{hokamp-liu-2017-lexically,
    title = "Lexically Constrained Decoding for Sequence Generation Using Grid Beam Search",
    author = "Hokamp, Chris  and
      Liu, Qun",
    booktitle = "Proceedings of the 55th Annual Meeting of the Association for Computational Linguistics (Volume 1: Long Papers)",
    month = jul,
    year = "2017",
    address = "Vancouver, Canada",
    publisher = "Association for Computational Linguistics",
    url = "https://aclanthology.org/P17-1141",
    doi = "10.18653/v1/P17-1141",
    pages = "1535--1546",
}

@ARTICLE{2022arXiv221109102V,
       author = {{Vilar}, David and {Freitag}, Markus and {Cherry}, Colin and {Luo}, Jiaming and {Ratnakar}, Viresh and {Foster}, George},
        title = "{Prompting PaLM for Translation: Assessing Strategies and Performance}",
      journal = {arXiv e-prints},
         year = 2022,
        month = nov,
          eid = {arXiv:2211.09102},
        pages = {arXiv:2211.09102},
          doi = {10.48550/arXiv.2211.09102},
archivePrefix = {arXiv},
       eprint = {2211.09102},
 primaryClass = {cs.CL},
       adsurl = {https://ui.adsabs.harvard.edu/abs/2022arXiv221109102V}
}

@ARTICLE{2022arXiv221202437A,
       author = {{Agrawal}, Sweta and {Zhou}, Chunting and {Lewis}, Mike and {Zettlemoyer}, Luke and {Ghazvininejad}, Marjan},
        title = "{In-context Examples Selection for Machine Translation}",
      journal = {arXiv e-prints},
         year = 2022,
        month = dec,
          eid = {arXiv:2212.02437},
        pages = {arXiv:2212.02437},
          doi = {10.48550/arXiv.2212.02437},
archivePrefix = {arXiv},
       eprint = {2212.02437},
 primaryClass = {cs.CL},
       adsurl = {https://ui.adsabs.harvard.edu/abs/2022arXiv221202437A}
}

@ARTICLE{2022arXiv220211822G,
       author = {{Garcia}, Xavier and {Firat}, Orhan},
        title = "{Using natural language prompts for machine translation}",
      journal = {arXiv e-prints},
         year = 2022,
        month = feb,
          eid = {arXiv:2202.11822},
        pages = {arXiv:2202.11822},
          doi = {10.48550/arXiv.2202.11822},
archivePrefix = {arXiv},
       eprint = {2202.11822},
 primaryClass = {cs.CL},
       adsurl = {https://ui.adsabs.harvard.edu/abs/2022arXiv220211822G}
}

@inproceedings{10.1145/3411763.3451760,
author = {Reynolds, Laria and McDonell, Kyle},
title = {Prompt Programming for Large Language Models: Beyond the Few-Shot Paradigm},
year = {2021},
isbn = {9781450380959},
publisher = {Association for Computing Machinery},
address = {New York, NY, USA},
url = {https://doi.org/10.1145/3411763.3451760},
doi = {10.1145/3411763.3451760},
booktitle = {Extended Abstracts of the 2021 CHI Conference on Human Factors in Computing Systems},
articleno = {314},
numpages = {7},
location = {Yokohama, Japan},
series = {CHI EA '21}
}

@ARTICLE{2022arXiv220501068Z,
       author = {{Zhang}, Susan and {Roller}, Stephen and {Goyal}, Naman and {Artetxe}, Mikel and {Chen}, Moya and {Chen}, Shuohui and {Dewan}, Christopher and {Diab}, Mona and {Li}, Xian and {Lin}, Xi Victoria and {Mihaylov}, Todor and {Ott}, Myle and {Shleifer}, Sam and {Shuster}, Kurt and {Simig}, Daniel and {Singh Koura}, Punit and {Sridhar}, Anjali and {Wang}, Tianlu and {Zettlemoyer}, Luke},
        title = "{OPT: Open Pre-trained Transformer Language Models}",
      journal = {arXiv e-prints},
         year = 2022,
        month = may,
          eid = {arXiv:2205.01068},
        pages = {arXiv:2205.01068},
          doi = {10.48550/arXiv.2205.01068},
archivePrefix = {arXiv},
       eprint = {2205.01068},
 primaryClass = {cs.CL},
       adsurl = {https://ui.adsabs.harvard.edu/abs/2022arXiv220501068Z}
}

@ARTICLE{2022arXiv221105100W,
       author = {{Le Scao}, Teven and {Fan}, Angela and {Akiki}, Christopher and {Pavlick}, Ellie and {Ili{\'c}}, Suzana and {Hesslow}, Daniel and {Castagn{\'e}}, Roman and {Sasha Luccioni}, Alexandra and {Yvon}, Fran{\c{c}}ois and {Gall{\'e}}, Matthias and {Tow}, Jonathan and {Rush}, Alexander M. and {Biderman}, Stella and {Webson}, Albert and {Sasanka Ammanamanchi}, Pawan and {Wang}, Thomas and {Sagot}, Beno{\^\i}t and {Muennighoff}, Niklas and {Villanova del Moral}, Albert and {Ruwase}, Olatunji and {Bawden}, Rachel and {Bekman}, Stas and {McMillan-Major}, Angelina and {Beltagy}, Iz and {Nguyen}, Huu and {Saulnier}, Lucile and {Tan}, Samson and {Ortiz Suarez}, Pedro and {Sanh}, Victor and {Lauren{\c{c}}on}, Hugo and {Jernite}, Yacine and {Launay}, Julien and {Mitchell}, Margaret and {Raffel}, Colin and {Gokaslan}, Aaron and {Simhi}, Adi and {Soroa}, Aitor and {Fikri Aji}, Alham and {Alfassy}, Amit and {Rogers}, Anna and {Kreisberg Nitzav}, Ariel and {Xu}, Canwen and {Mou}, Chenghao and {Emezue}, Chris and {Klamm}, Christopher and {Leong}, Colin and {van Strien}, Daniel and {Ifeoluwa Adelani}, David and {Radev}, Dragomir and {Gonz{\'a}lez Ponferrada}, Eduardo and {Levkovizh}, Efrat and {Kim}, Ethan and {Bar Natan}, Eyal and {De Toni}, Francesco and {Dupont}, G{\'e}rard and {Kruszewski}, Germ{\'a}n and {Pistilli}, Giada and {Elsahar}, Hady and {Benyamina}, Hamza and {Tran}, Hieu and {Yu}, Ian and {Abdulmumin}, Idris and {Johnson}, Isaac and {Gonzalez-Dios}, Itziar and {de la Rosa}, Javier and {Chim}, Jenny and {Dodge}, Jesse and {Zhu}, Jian and {Chang}, Jonathan and {Frohberg}, J{\"o}rg and {Tobing}, Joseph and {Bhattacharjee}, Joydeep and {Almubarak}, Khalid and {Chen}, Kimbo and {Lo}, Kyle and {Von Werra}, Leandro and {Weber}, Leon and {Phan}, Long and {Ben allal}, Loubna and {Tanguy}, Ludovic and {Dey}, Manan and {Romero Mu{\~n}oz}, Manuel and {Masoud}, Maraim and {Grandury}, Mar{\'\i}a and {{\v{S}}a{\v{s}}ko}, Mario and {Huang}, Max and {Coavoux}, Maximin and {Singh}, Mayank and {Tian-Jian Jiang}, Mike and {Chien Vu}, Minh and {Jauhar}, Mohammad A. and {Ghaleb}, Mustafa and {Subramani}, Nishant and {Kassner}, Nora and {Khamis}, Nurulaqilla and {Nguyen}, Olivier and {Espejel}, Omar and {de Gibert}, Ona and {Villegas}, Paulo and {Henderson}, Peter and {Colombo}, Pierre and {Amuok}, Priscilla and {Lhoest}, Quentin and {Harliman}, Rheza and {Bommasani}, Rishi and {L{\'o}pez}, Roberto Luis and {Ribeiro}, Rui and {Osei}, Salomey and {Pyysalo}, Sampo and {Nagel}, Sebastian and {Bose}, Shamik and {Muhammad}, Shamsuddeen Hassan and {Sharma}, Shanya and {Longpre}, Shayne and {Nikpoor}, Somaieh and {Silberberg}, Stanislav and {Pai}, Suhas and {Zink}, Sydney and {Timponi Torrent}, Tiago and {Schick}, Timo and {Thrush}, Tristan and {Danchev}, Valentin and {Nikoulina}, Vassilina and {Laippala}, Veronika and {Lepercq}, Violette and {Prabhu}, Vrinda and {Alyafeai}, Zaid and {Talat}, Zeerak and {Raja}, Arun and {Heinzerling}, Benjamin and {Si}, Chenglei and {Ta{\c{s}}ar}, Davut Emre and {Salesky}, Elizabeth and {Mielke}, Sabrina J. and {Lee}, Wilson Y. and {Sharma}, Abheesht and {Santilli}, Andrea and {Chaffin}, Antoine and {Stiegler}, Arnaud and {Datta}, Debajyoti and {Szczechla}, Eliza and {Chhablani}, Gunjan and {Wang}, Han and {Pandey}, Harshit and {Strobelt}, Hendrik and {Fries}, Jason Alan and {Rozen}, Jos and {Gao}, Leo and {Sutawika}, Lintang and {Saiful Bari}, M and {Al-shaibani}, Maged S. and {Manica}, Matteo and {Nayak}, Nihal and {Teehan}, Ryan and {Albanie}, Samuel and {Shen}, Sheng and {Ben-David}, Srulik and {Bach}, Stephen H. and {Kim}, Taewoon and {Bers}, Tali and {Fevry}, Thibault and {Neeraj}, Trishala and {Thakker}, Urmish and {Raunak}, Vikas and {Tang}, Xiangru and {Yong}, Zheng-Xin and {Sun}, Zhiqing and {Brody}, Shaked and {Uri}, Yallow and {Tojarieh}, Hadar and {Roberts}, Adam and {Chung}, Hyung Won and {Tae}, Jaesung and {Phang}, Jason and {Press}, Ofir and {Li}, Conglong and {Narayanan}, Deepak and {Bourfoune}, Hatim and {Casper}, Jared and {Rasley}, Jeff and {Ryabinin}, Max and {Mishra}, Mayank and {Zhang}, Minjia and {Shoeybi}, Mohammad and {Peyrounette}, Myriam and {Patry}, Nicolas and {Tazi}, Nouamane and {Sanseviero}, Omar and {von Platen}, Patrick and {Cornette}, Pierre and {Fran{\c{c}}ois Lavall{\'e}e}, Pierre and {Lacroix}, R{\'e}mi and {Rajbhandari}, Samyam and {Gandhi}, Sanchit and {Smith}, Shaden and {Requena}, St{\'e}phane and {Patil}, Suraj and {Dettmers}, Tim and {Baruwa}, Ahmed and {Singh}, Amanpreet and {Cheveleva}, Anastasia and {Ligozat}, Anne-Laure and {Subramonian}, Arjun and {N{\'e}v{\'e}ol}, Aur{\'e}lie and {Lovering}, Charles and {Garrette}, Dan and {Tunuguntla}, Deepak and {Reiter}, Ehud and {Taktasheva}, Ekaterina and {Voloshina}, Ekaterina and {Bogdanov}, Eli and {Indra Winata}, Genta and {Schoelkopf}, Hailey and {Kalo}, Jan-Christoph and {Novikova}, Jekaterina and {Zosa Forde}, Jessica and {Clive}, Jordan and {Kasai}, Jungo and {Kawamura}, Ken and {Hazan}, Liam and {Carpuat}, Marine and {Clinciu}, Miruna and {Kim}, Najoung and {Cheng}, Newton and {Serikov}, Oleg and {Antverg}, Omer and {van der Wal}, Oskar and {Zhang}, Rui and {Zhang}, Ruochen and {Gehrmann}, Sebastian and {Mirkin}, Shachar and {Pais}, Shani and {Shavrina}, Tatiana and {Scialom}, Thomas and {Yun}, Tian and {Limisiewicz}, Tomasz and {Rieser}, Verena and {Protasov}, Vitaly and {Mikhailov}, Vladislav and {Pruksachatkun}, Yada and {Belinkov}, Yonatan and {Bamberger}, Zachary and {Kasner}, Zden{\v{e}}k and {Rueda}, Alice and {Pestana}, Amanda and {Feizpour}, Amir and {Khan}, Ammar and {Faranak}, Amy and {Santos}, Ana and {Hevia}, Anthony and {Unldreaj}, Antigona and {Aghagol}, Arash and {Abdollahi}, Arezoo and {Tammour}, Aycha and {HajiHosseini}, Azadeh and {Behroozi}, Bahareh and {Ajibade}, Benjamin and {Saxena}, Bharat and {Mu{\~n}oz Ferrandis}, Carlos and {Contractor}, Danish and {Lansky}, David and {David}, Davis and {Kiela}, Douwe and {Nguyen}, Duong A. and {Tan}, Edward and {Baylor}, Emi and {Ozoani}, Ezinwanne and {Mirza}, Fatima and {Ononiwu}, Frankline and {Rezanejad}, Habib and {Jones}, Hessie and {Bhattacharya}, Indrani and {Solaiman}, Irene and {Sedenko}, Irina and {Nejadgholi}, Isar and {Passmore}, Jesse and {Seltzer}, Josh and {Bonis Sanz}, Julio and {Dutra}, Livia and {Samagaio}, Mairon and {Elbadri}, Maraim and {Mieskes}, Margot and {Gerchick}, Marissa and {Akinlolu}, Martha and {McKenna}, Michael and {Qiu}, Mike and {Ghauri}, Muhammed and {Burynok}, Mykola and {Abrar}, Nafis and {Rajani}, Nazneen and {Elkott}, Nour and {Fahmy}, Nour and {Samuel}, Olanrewaju and {An}, Ran and {Kromann}, Rasmus and {Hao}, Ryan and {Alizadeh}, Samira and {Shubber}, Sarmad and {Wang}, Silas and {Roy}, Sourav and {Viguier}, Sylvain and {Le}, Thanh and {Oyebade}, Tobi and {Le}, Trieu and {Yang}, Yoyo and {Nguyen}, Zach and {Kashyap}, Abhinav Ramesh and {Palasciano}, Alfredo and {Callahan}, Alison and {Shukla}, Anima and {Miranda-Escalada}, Antonio and {Singh}, Ayush and {Beilharz}, Benjamin and {Wang}, Bo and {Brito}, Caio and {Zhou}, Chenxi and {Jain}, Chirag and {Xu}, Chuxin and {Fourrier}, Cl{\'e}mentine and {Le{\'o}n Peri{\~n}{\'a}n}, Daniel and {Molano}, Daniel and {Yu}, Dian and {Manjavacas}, Enrique and {Barth}, Fabio and {Fuhrimann}, Florian and {Altay}, Gabriel and {Bayrak}, Giyaseddin and {Burns}, Gully and {Vrabec}, Helena U. and {Bello}, Imane and {Dash}, Ishani and {Kang}, Jihyun and {Giorgi}, John and {Golde}, Jonas and {Posada}, Jose David and {Rangasai Sivaraman}, Karthik and {Bulchandani}, Lokesh and {Liu}, Lu and {Shinzato}, Luisa and {Hahn de Bykhovetz}, Madeleine and {Takeuchi}, Maiko and {P{\`a}mies}, Marc and {Castillo}, Maria A and {Nezhurina}, Marianna and {S{\"a}nger}, Mario and {Samwald}, Matthias and {Cullan}, Michael and {Weinberg}, Michael and {De Wolf}, Michiel and {Mihaljcic}, Mina and {Liu}, Minna and {Freidank}, Moritz and {Kang}, Myungsun and {Seelam}, Natasha and {Dahlberg}, Nathan and {Michio Broad}, Nicholas and {Muellner}, Nikolaus and {Fung}, Pascale and {Haller}, Patrick and {Chandrasekhar}, Ramya and {Eisenberg}, Renata and {Martin}, Robert and {Canalli}, Rodrigo and {Su}, Rosaline and {Su}, Ruisi and {Cahyawijaya}, Samuel and {Garda}, Samuele and {Deshmukh}, Shlok S and {Mishra}, Shubhanshu and {Kiblawi}, Sid and {Ott}, Simon and {Sang-aroonsiri}, Sinee and {Kumar}, Srishti and {Schweter}, Stefan and {Bharati}, Sushil and {Laud}, Tanmay and {Gigant}, Th{\'e}o and {Kainuma}, Tomoya and {Kusa}, Wojciech and {Labrak}, Yanis and {Shailesh Bajaj}, Yash and {Venkatraman}, Yash and {Xu}, Yifan and {Xu}, Yingxin and {Xu}, Yu and {Tan}, Zhe and {Xie}, Zhongli and {Ye}, Zifan and {Bras}, Mathilde and {Belkada}, Younes and {Wolf}, Thomas},
        title = "{BLOOM: A 176B-Parameter Open-Access Multilingual Language Model}",
      journal = {arXiv e-prints},
         year = 2022,
        month = nov,
          eid = {arXiv:2211.05100},
        pages = {arXiv:2211.05100},
          doi = {10.48550/arXiv.2211.05100},
archivePrefix = {arXiv},
       eprint = {2211.05100},
 primaryClass = {cs.CL},
       adsurl = {https://ui.adsabs.harvard.edu/abs/2022arXiv221105100W}
}

@inproceedings{2021arXiv211210668L,
    title = "Few-shot Learning with Multilingual Generative Language Models",
    author = "Lin, Xi Victoria  and
      Mihaylov, Todor  and
      Artetxe, Mikel  and
      Wang, Tianlu  and
      Chen, Shuohui  and
      Simig, Daniel  and
      Ott, Myle  and
      Goyal, Naman  and
      Bhosale, Shruti  and
      Du, Jingfei  and
      Pasunuru, Ramakanth  and
      Shleifer, Sam  and
      Koura, Punit Singh  and
      Chaudhary, Vishrav  and
      O{'}Horo, Brian  and
      Wang, Jeff  and
      Zettlemoyer, Luke  and
      Kozareva, Zornitsa  and
      Diab, Mona  and
      Stoyanov, Veselin  and
      Li, Xian",
    booktitle = "Proceedings of the 2022 Conference on Empirical Methods in Natural Language Processing",
    month = dec,
    year = "2022",
    address = "Abu Dhabi, United Arab Emirates",
    publisher = "Association for Computational Linguistics",
    url = "https://aclanthology.org/2022.emnlp-main.616",
    pages = "9019--9052",
}

@inproceedings{NEURIPS2020_1457c0d6,
 author = {Brown, Tom and Mann, Benjamin and Ryder, Nick and Subbiah, Melanie and Kaplan, Jared D and Dhariwal, Prafulla and Neelakantan, Arvind and Shyam, Pranav and Sastry, Girish and Askell, Amanda and Agarwal, Sandhini and Herbert-Voss, Ariel and Krueger, Gretchen and Henighan, Tom and Child, Rewon and Ramesh, Aditya and Ziegler, Daniel and Wu, Jeffrey and Winter, Clemens and Hesse, Chris and Chen, Mark and Sigler, Eric and Litwin, Mateusz and Gray, Scott and Chess, Benjamin and Clark, Jack and Berner, Christopher and McCandlish, Sam and Radford, Alec and Sutskever, Ilya and Amodei, Dario},
 booktitle = {Advances in Neural Information Processing Systems},
 editor = {H. Larochelle and M. Ranzato and R. Hadsell and M.F. Balcan and H. Lin},
 pages = {1877--1901},
 publisher = {Curran Associates, Inc.},
 title = {Language Models are Few-Shot Learners},
 url = {https://proceedings.neurips.cc/paper_files/paper/2020/file/1457c0d6bfcb4967418bfb8ac142f64a-Paper.pdf},
 volume = {33},
 year = {2020}
}

@ARTICLE{2023arXiv230304048W,
       author = {{Wang}, Jiaan and {Liang}, Yunlong and {Meng}, Fandong and {Shi}, Haoxiang and {Li}, Zhixu and {Xu}, Jinan and {Qu}, Jianfeng and {Zhou}, Jie},
        title = "{Is ChatGPT a Good NLG Evaluator? A Preliminary Study}",
      journal = {arXiv e-prints},
         year = 2023,
        month = mar,
          eid = {arXiv:2303.04048},
        pages = {arXiv:2303.04048},
          doi = {10.48550/arXiv.2303.04048},
archivePrefix = {arXiv},
       eprint = {2303.04048},
 primaryClass = {cs.CL},
       adsurl = {https://ui.adsabs.harvard.edu/abs/2023arXiv230304048W}
}

@article{nllb2022,
  author    = {NLLB-Team},
  title     = {No Language Left Behind: Scaling Human-Centered Machine Translation},
  year      = {2022}
}

@ARTICLE{2023arXiv230506575L,
       author = {{Lu}, Hongyuan and {Yang}, Haoran and {Huang}, Haoyang and {Zhang}, Dongdong and {Lam}, Wai and {Wei}, Furu},
        title = "{Chain-of-Dictionary Prompting Elicits Translation in Large Language Models}",
      journal = {arXiv e-prints},
         year = 2023,
        month = may,
          eid = {arXiv:2305.06575},
        pages = {arXiv:2305.06575},
          doi = {10.48550/arXiv.2305.06575},
archivePrefix = {arXiv},
       eprint = {2305.06575},
 primaryClass = {cs.CL},
       adsurl = {https://ui.adsabs.harvard.edu/abs/2023arXiv230506575L}
}

@inproceedings{wang-etal-2023-towards,
    title = "Towards Understanding Chain-of-Thought Prompting: An Empirical Study of What Matters",
    author = "Wang, Boshi  and
      Min, Sewon  and
      Deng, Xiang  and
      Shen, Jiaming  and
      Wu, You  and
      Zettlemoyer, Luke  and
      Sun, Huan",
    editor = "Rogers, Anna  and
      Boyd-Graber, Jordan  and
      Okazaki, Naoaki",
    booktitle = "Proceedings of the 61st Annual Meeting of the Association for Computational Linguistics (Volume 1: Long Papers)",
    month = jul,
    year = "2023",
    address = "Toronto, Canada",
    publisher = "Association for Computational Linguistics",
    url = "https://aclanthology.org/2023.acl-long.153",
    doi = "10.18653/v1/2023.acl-long.153",
    pages = "2717--2739",
}

@inproceedings{10.5555/3600270.3602070,
author = {Wei, Jason and Wang, Xuezhi and Schuurmans, Dale and Bosma, Maarten and Ichter, Brian and Xia, Fei and Chi, Ed H. and Le, Quoc V. and Zhou, Denny},
title = {Chain-of-thought prompting elicits reasoning in large language models},
year = {2024},
isbn = {9781713871088},
publisher = {Curran Associates Inc.},
address = {Red Hook, NY, USA},
booktitle = {Proceedings of the 36th International Conference on Neural Information Processing Systems},
articleno = {1800},
numpages = {14},
location = {New Orleans, LA, USA},
series = {NIPS '22}
}

@inproceedings{zhu-etal-2024-multilingual,
    title = "Multilingual Machine Translation with Large Language Models: Empirical Results and Analysis",
    author = "Zhu, Wenhao  and
      Liu, Hongyi  and
      Dong, Qingxiu  and
      Xu, Jingjing  and
      Huang, Shujian  and
      Kong, Lingpeng  and
      Chen, Jiajun  and
      Li, Lei",
    editor = "Duh, Kevin  and
      Gomez, Helena  and
      Bethard, Steven",
    booktitle = "Findings of the Association for Computational Linguistics: NAACL 2024",
    month = jun,
    year = "2024",
    address = "Mexico City, Mexico",
    publisher = "Association for Computational Linguistics",
    url = "https://aclanthology.org/2024.findings-naacl.176",
    doi = "10.18653/v1/2024.findings-naacl.176",
    pages = "2765--2781",
}

@inproceedings{li-etal-2023-codeie,
    title = "{C}ode{IE}: Large Code Generation Models are Better Few-Shot Information Extractors",
    author = "Li, Peng  and
      Sun, Tianxiang  and
      Tang, Qiong  and
      Yan, Hang  and
      Wu, Yuanbin  and
      Huang, Xuanjing  and
      Qiu, Xipeng",
    editor = "Rogers, Anna  and
      Boyd-Graber, Jordan  and
      Okazaki, Naoaki",
    booktitle = "Proceedings of the 61st Annual Meeting of the Association for Computational Linguistics (Volume 1: Long Papers)",
    month = jul,
    year = "2023",
    address = "Toronto, Canada",
    publisher = "Association for Computational Linguistics",
    url = "https://aclanthology.org/2023.acl-long.855",
    doi = "10.18653/v1/2023.acl-long.855",
    pages = "15339--15353",
}

@inproceedings{zhang-etal-2023-self,
    title = "Self-Edit: Fault-Aware Code Editor for Code Generation",
    author = "Zhang, Kechi  and
      Li, Zhuo  and
      Li, Jia  and
      Li, Ge  and
      Jin, Zhi",
    editor = "Rogers, Anna  and
      Boyd-Graber, Jordan  and
      Okazaki, Naoaki",
    booktitle = "Proceedings of the 61st Annual Meeting of the Association for Computational Linguistics (Volume 1: Long Papers)",
    month = jul,
    year = "2023",
    address = "Toronto, Canada",
    publisher = "Association for Computational Linguistics",
    url = "https://aclanthology.org/2023.acl-long.45",
    doi = "10.18653/v1/2023.acl-long.45",
    pages = "769--787",
}

@inproceedings{
hu2024chainofsymbol,
title={Chain-of-Symbol Prompting For Spatial Reasoning in Large Language Models},
author={Hanxu Hu and Hongyuan Lu and Huajian Zhang and Yun-Ze Song and Wai Lam and Yue Zhang},
booktitle={First Conference on Language Modeling},
year={2024},
url={https://openreview.net/forum?id=Hvq9RtSoHG}
}

@inproceedings{hou-etal-2025-lne,
    title = "{LNE}-Blocking: An Efficient Framework for Contamination Mitigation Evaluation on Large Language Models",
    author = "Hou, Ruijie  and
      Jiao, Yueyang  and
      Hu, Hanxu  and
      Li, Yingming  and
      Lam, Wai  and
      Zhang, Huajian  and
      Lu, Hongyuan",
    editor = "Christodoulopoulos, Christos  and
      Chakraborty, Tanmoy  and
      Rose, Carolyn  and
      Peng, Violet",
    booktitle = "Findings of the Association for Computational Linguistics: EMNLP 2025",
    month = nov,
    year = "2025",
    address = "Suzhou, China",
    publisher = "Association for Computational Linguistics",
    url = "https://aclanthology.org/2025.findings-emnlp.188/",
    doi = "10.18653/v1/2025.findings-emnlp.188",
    pages = "3512--3528",
    ISBN = "979-8-89176-335-7"
}
